\pdfoutput=1

\documentclass[11pt]{article}

\usepackage[]{ACL2023}

\usepackage{times}
\usepackage{latexsym}

\usepackage{amssymb,mathtools}
\usepackage{dsfont}
\usepackage{graphicx,paralist,multirow}
\usepackage{makecell,booktabs}
\usepackage{cleveref}
\crefformat{section}{\S#2#1#3}
\crefformat{subsection}{\S#2#1#3}
\usepackage{xcolor,colortbl}
\usepackage{subcaption}

\usepackage[T1]{fontenc}

\usepackage[utf8]{inputenc}

\usepackage{microtype}

\usepackage{inconsolata}
\newcommand{\good}{\emph{stable}}
\newcommand{\Dtrain}[1]{D_{\text{tr}}#1}
\newcommand{\Ddev}[1]{D_{\text{dev}}#1}
\newcommand{\Dtest}[1]{D_{\text{test}}#1}
\newcommand{\Dicl}[1]{\mathcal{D_{\text{ICL}}}#1}
\newcommand{\Prompt}[1]{\mathcal{Z}#1}
\newcommand{\DMsubset}{\mathcal{S}}
\newcommand{\DevInp}{\bar{x}}
\newcommand{\DevOut}{\bar{y}}
\newcommand{\DevEx}{(\DevInp, \DevOut)}
\newcommand{\Ntrain}{N_{\text{tr}}}
\newcommand{\ssim}{\boldsymbol{s}_{ca}}
\newcommand{\sshap}{\boldsymbol{s}_{shap}}
\newcommand{\zdm}{z}
\newcommand{\Simple}{\textsc{CondAcc}}
\newcommand{\Datamodels}{\textsc{Datamodels}}
\newcommand{\RandAll}{\textsc{All}}
\newcommand{\RandSub}{\textsc{Random}}
\newcommand{\Best}{\textsc{TopPrompts}}

\newcommand{\One}{\textsc{OneShot}}
\newcommand{\UnSimple}{\textsc{Un-CondAcc}}
\newcommand{\UnOne}{\textsc{Un-OneShot}}
\newcommand{\UnBest}{\textsc{Un-TopPrompts}}
\newcommand{\UnAll}{\textsc{Un-All}}
\newcommand{\Calib}{\textsc{Calib}}
\newcommand{\MaxShot}{\textsc{MaxShot}}

\title{Data Curation Alone Can Stabilize In-context Learning}

\author{Ting-Yun Chang \and Robin Jia \\
  University of Southern California \\
  \texttt{\{tingyun,robinjia\}@usc.edu} \\
}

\begin{document}

\maketitle
\begin{abstract}
In-context learning (ICL) enables large language models (LLMs) to perform new tasks by prompting them with a sequence of training examples. However, it is known that ICL is very sensitive to the choice of training examples: randomly sampling examples from a training set leads to high variance in performance. In this paper, we show that carefully curating a subset of training data greatly \emph{stabilizes} ICL performance without any other changes to the ICL algorithm (e.g., prompt retrieval or calibration). We introduce two methods to choose training subsets---both score training examples individually, then select the highest-scoring ones. \Simple\ scores a training example by its average dev-set ICL accuracy when combined with random training examples, while \Datamodels\ learns linear regressors that estimate how the presence of each training example influences LLM outputs. 
Across five tasks and two LLMs, sampling from \good\ subsets selected by \Simple\ and \Datamodels\ improves average accuracy over sampling from the entire training set by 7.7\% and 6.3\%, respectively.
Surprisingly, the \good\ subset examples are not especially diverse in content or low in perplexity, in contrast with other work suggesting that diversity and perplexity are important when prompting LLMs. 

\end{abstract}

\section{Introduction}
\label{intro}
In-context learning (ICL) is a new paradigm for few-shot learning with pretrained large language models (LLMs) without any parameter updates.
In ICL, an LLM can perform a new task simply by conditioning on a prompt\footnote{Prior work has different definitions of \emph{prompt}; in this paper, we fix the task templates and follow \citet{rubin2021learning} to denote \emph{prompt} as a sequence of training examples for ICL.} consisting of a sequence of labeled training examples.
First introduced by GPT-3~\cite{brown2020language}, ICL with LLMs has reached state-of-the-art few-shot performance across many tasks~\cite{rae2021scaling,smith2022using,thoppilan2022lamda,chowdhery2022palm}.
Compared with alternatives that use fine-tuning~\cite{devlin2018bert,schick2020exploiting,gao2020making}, ICL does not require task-specific training, which enables its use with very large language models, 
and it uses a unified model for all tasks, enabling easier deployment.

\begin{figure}[t!]
  \centering
  \includegraphics[width=0.9\linewidth]{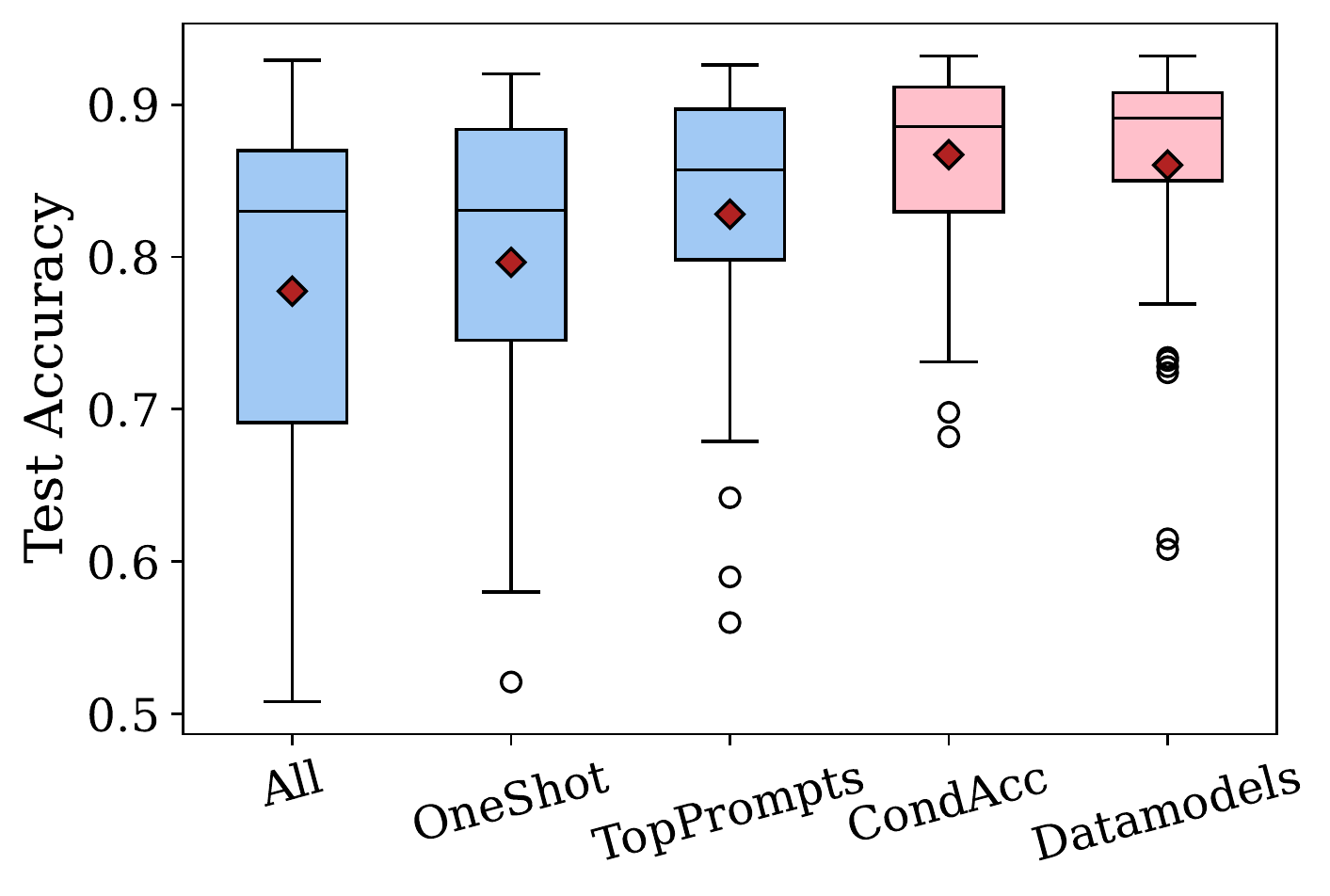}
  \caption{4-shot ICL performance of GPTJ on SST2. Each boxplot summarizes the results of 50 sampled prompts. Compared with baselines (blue), our methods (pink) can greatly stablilize performance, having higher average accuracy (red diamonds) and lower variance.}
  \label{fig:fig1}
\end{figure}
Despite its impressive few-shot performance, ICL often exhibits unintuitive behavior~\cite{min2022rethinking}.
The standard ICL approach is to randomly sample a few examples from a training set to construct a prompt~\cite{brown2020language}; however, prior work~\cite{liu2021makes,zhao2021calibrate,lu2021fantastically} has found that ICL is very sensitive to the choice of training examples and their order in the prompt.
ICL is also sensitive to small changes in prompt format~\cite{chen2022relation}.

In this paper, we show that carefully curating a smaller training dataset from a larger pool can make ICL much more stable.
We define a training subset $\mathcal{E}$ to be \good\ if randomly sampling a sequence of examples as a prompt from $\mathcal{E}$ yields much higher average and worst-case accuracy than randomly sampling from the original training set.
We propose two methods to identify such a stable subset.
Our \Simple\ method scores a training example by its average dev-set ICL accuracy when combined with random training examples; these scores are closely related to Data Shapley values~\cite{ghorbani2019data}.
Our \Datamodels\ method fits a linear regressor that predicts the LLM's output based on which example is present at each index in the prompt;
we score a training example highly if the associated weights from the linear model indicate that its presence improves accuracy.
For both methods, we then select training examples with the highest scores to form the stable subset.
While some prior work improves ICL accuracy by retrieving a suitable prompt for each test example~\cite{liu2021makes,rubin2021learning,su2022selective}, we show that it is possible to achieve stably good accuracy with a randomly sampled prompt for all test examples, when given the ``right'' training (sub)set.

Our subset selection methods greatly improve performance across 5 classification datasets and 4 LLMs, with main experiments on GPTJ-6B~\cite{gpt-j} and OPT-13B~\cite{zhang2022opt}.
On average, \Simple\ and \Datamodels\ outperform the baseline that uses the entire training set without selection (named \RandAll) by 7.7\% and 6.3\%, respectively, when comparing the average accuracy over multiple sampled prompts.
In contrast, baselines that choose examples found in high-performing prompts (\Best) or examples that lead to high one-shot ICL accuracy (\One) to form the subsets do not perform as well (see Figure~\ref{fig:fig1}).
Our stable subset examples generalize to out-of-distribution test data, and 
we can even find stable subsets for binary classification tasks that only contain examples of one label;
these findings suggest that the stable subset examples help LLMs understand the overall task definition.

Finally, we study what makes stable subset examples special by analyzing sequence length, perplexity, and diversity in both raw text and embedding spaces.
We find that these examples do not have abnormally long sequence lengths or high perplexities.
In contrast with prior work optimizing diversity for prompt selection~\cite{su2022selective,Ye2022ComplementaryEF}, we find our stable subsets no more diverse than random subsets of the training data. 
In summary, we show that curating training data appropriately leads to more stable and accurate ICL performance; we hope future work can develop new strategies for writing such helpful examples.
Code and data are publicly available at \url{https://github.com/terarachang/DataICL}.

\begin{figure*}[t!]
  \centering
  \includegraphics[width=0.95\linewidth]{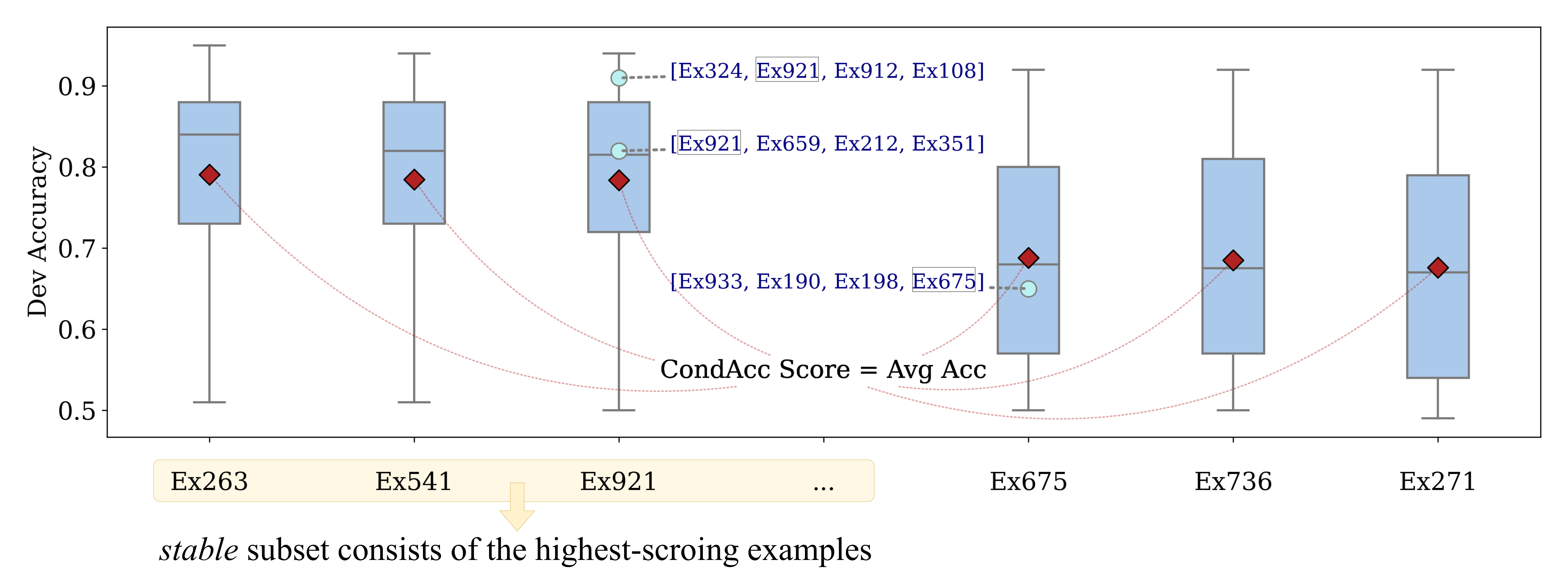}
  \caption{An overview of our \Simple\ method, which scores training examples individually using its average accuracy (red diamonds) when combined with other random training examples. Each boxplot summarizes the dev-set accuracies conditioned on a training example appearing in the sampled prompts.}
  \label{fig:main}
\end{figure*}
\section{Problem Setups}
\label{sec:problem_setup}
We use the original ICL formulation proposed by GPT-3, also known as the \emph{direct} method, for all our experiments. 
Specifically, an LLM performs in-context learning on a new task based on a task-specific prompt $\Prompt$ formed by concatenating $K$ labeled training examples, i.e., $\Prompt = [z_1,...,z_K]$, where each $z_j$ is a training example $(x, y)$ consisting of an input $x$ and label $y$.
The LLM then makes predictions on a test input $x_{\text{test}}$ conditioned on the prompt $\Prompt$ 
 followed by $x_{\text{test}}$, denoted by
${\arg\max}_{y\in \mathcal{C}} P(y|\Prompt, x_{\text{test}})$, where $\mathcal{C}$ is the set of possible labels.

Given a training set $\Dtrain$, a dev set $\Ddev$, a held-out test set $\Dtest$, and a predefined number of shots $K$, our goal is to select a stable training subset $\mathcal{E} \subset \Dtrain, |\mathcal{E}| > K$, such that randomly sampling a sequence of $K$ examples from $\mathcal{E}$ to form a prompt generally yields good performance on $\Dtest$.

We propose two setups, \textbf{Labeled} and \textbf{Unlabeled}.
In \textbf{Labeled}, our goal is to study which training examples consistently lead to high ICL accuracy. We assume access to a large labeled $\Dtrain^{L} = \{(x_i, y_i)\}_{i=1}^{\Ntrain}$ of input-label pair, and a small labeled $\Ddev$. 
\textbf{Unlabeled} is closer to the true few-shot learning setup~\cite{perez2021true}, where we only have access to $\Ddev$ and a large unlabeled training set $\{x_i\}_{i=1}^{\Ntrain}$.
We pair each input $x_i$ with every possible label $y \in \mathcal{C}$ to create an unlabeled training set $\Dtrain^{U} = \{\{(x_i,\tilde{y})|\tilde{y}\in\mathcal{C}\}\}_{i=1}^{\Ntrain}$.
In both setups, our goal is to select a subset of $E$ training examples from either $\Dtrain^{L}$ or $\Dtrain^{U}$. Note that we may select examples with incorrect labels under \textbf{Unlabeled}.
We only use the large labeled $\Dtest$ for evaluating our selection methods.

\section{Methods}
\label{sec:methods}
We propose the following steps to identify a stable subset of $E$ training examples:
\begin{enumerate}
  \item Construct $\Dicl$, a large set of $M$ prompts, each consisting of $K$ examples randomly sampled from $\Dtrain$. We then run ICL on the dev set $M$ times given different prompts.
  \item Estimate the value of each training example based on the results of Step 1. We propose two methods to do this: \Simple\ (\cref{sec:simple}) and \Datamodels\ (\cref{sec:datamodel}).
  \item Select training examples with the highest importance scores per class to make up $\mathcal{E}$ (\cref{sec:step3}).
\end{enumerate}

\subsection{CondAcc}
\label{sec:simple}
We hypothesize that a good training example leads to higher accuracy, on average, when occurs in a prompt.
Given a prompt $\Prompt$, we denote its dev set ICL accuracy as $\text{Acc}(\Prompt)$.
Thus, a simple way to score the $i$-th training example $(x_i, y_i) \in \Dtrain $ is to calculate the expected accuracy conditioned on this example appearing in a prompt $\Prompt$ from $\Dicl$:
\begin{equation}
    \ssim(i) = \mathbb{E}_{\Prompt \sim \Dicl\ } \left[ \text{Acc}(\Prompt) \mid (x_i, y_i) \in \Prompt \right] 
\label{eq:simple_score}
\end{equation}
We ensure that each training example occurs in $\Dicl$ multiple times in different orders and with different examples.

In Appendix~\ref{app:shapley}, we show that Eq.~\ref{eq:simple_score} is similar to Data Shapley value~\cite{ghorbani2019data}, where we define the valuation function of Data Shapley on subsets of $K$ training examples since we focus on $K$-shot ICL.

\subsection{Datamodels}
\label{sec:datamodel}
The \Simple\ method does not consider the order of training examples, which has a great impact on ICL performance.
Also, it takes a simple average over the dev set, ignoring the LLM's confidence in individual dev examples.
Thus, we propose another data valuation method that leverages Datamodels~\cite{ilyas2022datamodels} for ICL.

\citet{ilyas2022datamodels} study a complex target model's behavior in terms
of the training data by replacing it with a linear, easy-to-analyze, proxy model called a \emph{datamodel}.
Specifically, given a subset of training data $\DMsubset$, a datamodel predicts the \emph{outcome} of a test input $\DevInp$ if training the target model on $\DMsubset$. 
The \emph{outcome} $f(\DevInp; \DMsubset)$ is defined as the margin of the correct class, i.e., the logit for the correct class minus the highest incorrect logit.
To train datamodels, ~\citet{ilyas2022datamodels} first create a dataset consisting of subset-outcome pairs $(\DMsubset, f(\DevInp; \DMsubset))$, requiring training the target model from scratch multiple times on different subsets to obtain the outcomes.
In our work, the target model is a pretrained LLM, and is inference-only during ICL.
Thus, our dataset collection only requires inference of the LLM multiple times with different prompts in $\Dicl$.

In particular, given a prompt $\Prompt = [z_1,...,z_K]$, we run ICL on a dev example $(\DevInp, \DevOut)$ and define an LLM's \emph{outcome} as $f(\DevInp; \Prompt) = o(\DevOut | \Prompt, \DevInp) - \max_{y^{\prime}\in\mathcal{C},\, y^{\prime} \ne \DevOut} o(y^{\prime} | \Prompt, \DevInp)$, where $o(y | \Prompt, \DevInp)$ is the output logit of the LLM on label $y$ before softmax.
We hypothesize that we can approximate an LLM's \emph{outcome} with a linear regressor taking two simple features: the existence of a training example and its index in $\Prompt$, which we consider to be the most important factors in ICL.
Our linear datamodel is parameterized by weights $w \in \mathbb{R}^{\Ntrain \times K}$ and bias $b \in \mathbb{R}$;
we use $w(i,j) \in \mathbb{R}$ to denote the weight for the $i$-th training example appearing at index $j$.
For each dev example $\DevEx \in \Ddev$, we train a datamodel $g_{w,b}$ to predict the LLM's \emph{outcome} of $\DevInp$, $f(\DevInp; \Prompt) \in \mathbb{R}$, with mean-squared error:
\begin{align} 
g_{w,b}(\DevInp; \Prompt) &= \sum_{j=1}^{K} w(\operatorname{id}(\zdm_j),j) + b, \\
\underset{w,b}\min \;\frac{1}{M}\sum_{n=1}^{M} \; &  (g_{w,b}(\DevInp; \Prompt_n) - f(\DevInp; \Prompt_n))^2,
\end{align}
where $\zdm_j$ is the training example at index $j$ in the prompt $\Prompt$, and $\operatorname{id}(\cdot)$ maps $\zdm_j$ back to its example ID $i$ in the training set, i.e., $\zdm_j = (x_i, y_i)$.

By definition, $f(\DevInp; \Prompt) > 0$ if the LLM is correct on $\DevInp$.
Hence, a positive $w(i,j)$ indicates that having the $i$-th training example at index $j$ in the prompt encourages answering correctly.
We hypothesize that a good training example should have a beneficial effect for many dev examples, regardless of its index in the prompts.
Thus, we can aggregate the datamodels of all dev examples and marginalize over all possible orders to calculate the score of the $i$-th training example:
\begin{equation}
\label{eq:datamodel_score}
    \boldsymbol{s}_{dm}(i) = \sum_{\DevEx \in \Ddev\ }\sum_{j=1}^{K}\mathds{1} \{w_{\DevInp}(i,j) > 0\},
\end{equation}
where $w_{\DevInp}(i,j)$ is the weight value $w(i,j)$ of the datamodel for $\DevInp$.
Calculating the total number of positive weights empirically works better than averaging the weights of all the datamodels.

In Appendix~\ref{app:dm_test}, we validate our hypothesis that we can linearize an LLM's \emph{outcomes} with two simple features.
Table~\ref{table:dm_test} shows that our datamodels can accurately approximate the outcomes of unseen prompts outside of $\Dicl$ across different tasks.
\subsection{Select Training Examples}
\label{sec:step3}
Now that we assign a score for each training example (Eq.~\ref{eq:simple_score}, \ref{eq:datamodel_score}), let $C$ be the number of classes and $E' = E / C$, we select the top-$E'$ training examples of each class with the highest scores to form the stable subset $\mathcal{E} \subset \Dtrain, |\mathcal{E}| = E$.

\begin{table*}[!t]
\begin{center}
\centering
\resizebox{2.0\columnwidth}{!}
{
\begin{tabular}{lcccccccccc>{\columncolor[gray]{0.9}}c}
\toprule
      & \multicolumn{2}{c}{\textbf{SST-2}} & \multicolumn{2}{c}{\textbf{BoolQ}} & \multicolumn{2}{c}{\textbf{Subj}} & \multicolumn{2}{c}{\textbf{Scicite}} & \multicolumn{2}{c}{\textbf{AGNews}} & \cellcolor{white}\\

\cmidrule(lr){2-3}
\cmidrule(lr){4-5}
\cmidrule(lr){6-7}
\cmidrule(lr){8-9}
\cmidrule(lr){10-11}
& Avg std & Min & Avg std & Min & Avg std & Min & Avg std & Min & Avg std & Min & \multirow{-2}{*}{\cellcolor{white}
 \makecell{ Avg. \\ Tasks} }\\
\midrule
\emph{GPTJ-6B}\\
~  \RandAll & $77.8_{\hspace{0.05cm}11.2}$ & $50.8$ & $61.0_{\hspace{0.05cm}3.8}$ & $49.7$ & $59.8_{\hspace{0.05cm}8.3}$ & $50.1$ & $43.8_{\hspace{0.05cm}7.2}$ & $33.6$ & $83.5_{\hspace{0.05cm}3.8}$ & $70.4$ & $65.2$\\
~~  + \Calib & $75.5_{\hspace{0.05cm}9.5}$ & $53.6$ & $61.2_{\hspace{0.05cm}3.9}$ & $50.4$ & $70.4_{\hspace{0.05cm}7.7}$ & $\textbf{55.7}$ & $35.4_{\hspace{0.05cm}2.6}$ & $32.8$ & $85.2_{\hspace{0.05cm}2.7}$ & $78.0$ & $65.5$\\
~  \RandSub & $74.6_{\hspace{0.05cm}11.4}$ & $50.3$ & $60.0_{\hspace{0.05cm}4.3}$ & $49.5$ & $59.9_{\hspace{0.05cm}10.4}$ & $50.1$ & $46.4_{\hspace{0.05cm}6.9}$ & $35.5$ & $82.5_{\hspace{0.05cm}4.7}$ & $67.1$ & $64.7$\\
~  \One & $79.6_{\hspace{0.05cm}10.5}$ & $52.1$ & $63.8_{\hspace{0.05cm}2.7}$ & $56.4$ & $63.3_{\hspace{0.05cm}10.1}$ & $50.1$ & $44.8_{\hspace{0.05cm}5.9}$ & $33.8$ & $83.3_{\hspace{0.05cm}3.4}$ & $71.9$ & $67.0$\\
~  \Best-5 & $82.8_{\hspace{0.05cm}8.6}$ & $56.0$ & $62.3_{\hspace{0.05cm}3.0}$ & $54.3$ & $65.5_{\hspace{0.05cm}9.7}$ & $50.1$ & $50.4_{\hspace{0.05cm}6.0}$ & $36.9$ & $84.4_{\hspace{0.05cm}3.3}$ & $74.3$ & $69.1$\\
~  \Best-10 & $78.5_{\hspace{0.05cm}9.3}$ & $52.4$ & $61.2_{\hspace{0.05cm}4.0}$ & $51.1$ & $65.1_{\hspace{0.05cm}10.7}$ & $50.1$ & $49.4_{\hspace{0.05cm}5.5}$ & $36.2$ & $85.4_{\hspace{0.05cm}2.4}$ & $76.3$ & $67.9$\\
~  \cellcolor{blue!25}\Simple & $\textbf{86.7}_{\hspace{0.05cm}5.9}$ & $\textbf{68.2}$ & $65.1_{\hspace{0.05cm}1.6}$ & $61.1$ & $\textbf{70.5}_{\hspace{0.05cm}10.4}$ & $50.2$ & $52.3_{\hspace{0.05cm}4.4}$ & $42.0$ & $\textbf{87.3}_{\hspace{0.05cm}2.6}$ & $70.5$ & $\textbf{72.4}$\\
~  \cellcolor{blue!25}\Datamodels & $86.0_{\hspace{0.05cm}7.5}$ & $60.8$ & $\textbf{65.2}_{\hspace{0.05cm}0.9}$ & $\textbf{63.4}$ & $69.4_{\hspace{0.05cm}10.7}$ & $50.4$ & $\textbf{54.5}_{\hspace{0.05cm}3.8}$ & $\textbf{43.9}$ & $86.9_{\hspace{0.05cm}1.4}$ & $\textbf{82.8}$ & $\textbf{72.4}$\\
\midrule
~  \UnAll & $71.0_{\hspace{0.05cm}11.9}$ & $50.0$ & $60.8_{\hspace{0.05cm}3.5}$ & $49.6$ & $60.1_{\hspace{0.05cm}8.8}$ & $50.1$ & $42.0_{\hspace{0.05cm}7.0}$ & $33.5$ & $75.1_{\hspace{0.05cm}9.9}$ & $46.5$ & 61.8\\
~  \UnOne & $81.9_{\hspace{0.05cm}6.3}$ & $\textbf{68.5}$ & $62.6_{\hspace{0.05cm}3.3}$ & $55.6$ & $61.0_{\hspace{0.05cm}8.7}$ & $50.1$ & $43.5_{\hspace{0.05cm}6.7}$ & $33.4$ & $78.1_{\hspace{0.05cm}4.2}$ & $69.8$ & 65.4\\
~  \UnBest-5 & $80.1_{\hspace{0.05cm}10.5}$ & $56.8$ & $61.2_{\hspace{0.05cm}3.3}$ & $51.9$ & $60.7_{\hspace{0.05cm}10.0}$ & $50.1$ & $48.7_{\hspace{0.05cm}6.9}$ & $33.0$ & $76.4_{\hspace{0.05cm}9.8}$ & $53.0$ & 65.4\\
~  \cellcolor{blue!25}\UnSimple & $\textbf{85.3}_{\hspace{0.05cm}6.8}$ & $60.5$ & $\textbf{63.7}_{\hspace{0.05cm}2.2}$ & $\textbf{56.0}$ & $\textbf{66.0}_{\hspace{0.05cm}10.6}$ & $50.1$ & $\textbf{54.2}_{\hspace{0.05cm}3.4}$ & $\textbf{45.9}$ & $\textbf{87.1}_{\hspace{0.05cm}1.1}$ & $\textbf{84.6}$ & $\textbf{71.3}$\\

\midrule
\midrule

\emph{OPT-13B}\\
~  \RandAll & $68.5_{\hspace{0.05cm}14.0}$ & $50.0$ & $65.2_{\hspace{0.05cm}5.6}$ & $49.7$ & $60.9_{\hspace{0.05cm}10.2}$ & $49.8$ & $42.8_{\hspace{0.05cm}3.6}$ & $35.0$ & $81.6_{\hspace{0.05cm}5.9}$ & $64.2$ & $63.8$\\
~~ + \Calib & $\textbf{84.7}_{\hspace{0.05cm}6.8}$ & $51.7$ & $65.5_{\hspace{0.05cm}4.9}$ & $51.8$ & $63.7_{\hspace{0.05cm}8.9}$ & $47.9$ & $35.5_{\hspace{0.05cm}1.8}$ & $31.2$ & $81.8_{\hspace{0.05cm}4.1}$ & $70.7$ & $66.2$\\
~  \RandSub & $67.7_{\hspace{0.05cm}14.1}$ & $50.0$ & $64.7_{\hspace{0.05cm}6.4}$ & $49.3$ & $61.2_{\hspace{0.05cm}9.5}$ & $49.9$ & $41.2_{\hspace{0.05cm}4.6}$ & $33.3$ & $78.0_{\hspace{0.05cm}7.5}$ & $61.4$ & $62.6$\\
~  \One & $75.6_{\hspace{0.05cm}13.1}$ & $50.7$ & $68.3_{\hspace{0.05cm}2.3}$ & $62.7$ & $60.5_{\hspace{0.05cm}9.9}$ & $49.9$ & $41.9_{\hspace{0.05cm}3.8}$ & $33.4$ & $84.2_{\hspace{0.05cm}2.9}$ & $73.1$ & $66.1$\\
~  \Best-5 & $69.6_{\hspace{0.05cm}14.7}$ & $50.0$ & $63.5_{\hspace{0.05cm}6.3}$ & $51.0$ & $67.4_{\hspace{0.05cm}12.7}$ & $50.0$ & $45.9_{\hspace{0.05cm}4.3}$ & $36.0$ & $83.9_{\hspace{0.05cm}3.1}$ & $74.0$ & $66.1$\\
~  \Best-10 & $72.9_{\hspace{0.05cm}15.6}$ & $50.0$ & $65.5_{\hspace{0.05cm}5.2}$ & $50.4$ & $68.5_{\hspace{0.05cm}13.4}$ & $49.9$ & $44.6_{\hspace{0.05cm}3.9}$ & $36.7$ & $84.4_{\hspace{0.05cm}3.5}$ & $70.9$ & $67.2$\\
~  \cellcolor{blue!25}\Simple & $83.6_{\hspace{0.05cm}9.1}$ & $56.1$ & $\textbf{69.4}_{\hspace{0.05cm}2.1}$ & $\textbf{62.8}$ & $\textbf{70.6}_{\hspace{0.05cm}11.9}$ & $50.0$ & $\textbf{49.4}_{\hspace{0.05cm}3.3}$ & $\textbf{41.1}$ & $\textbf{87.0}_{\hspace{0.05cm}1.0}$ & $\textbf{83.6}$ & $\textbf{72.0}$\\
~  \cellcolor{blue!25}\Datamodels & $81.3_{\hspace{0.05cm}10.3}$ & $\textbf{60.3}$ & $69.3_{\hspace{0.05cm}3.8}$ & $57.3$ & $63.0_{\hspace{0.05cm}9.4}$ & $\textbf{50.1}$ & $46.3_{\hspace{0.05cm}3.9}$ & $37.4$ & $85.7_{\hspace{0.05cm}1.7}$ & $81.8$ & $69.1$\\
\midrule
~  \UnAll & $61.6_{\hspace{0.05cm}13.6}$ & $50.0$ & $64.8_{\hspace{0.05cm}5.3}$ & $49.3$ & $55.8_{\hspace{0.05cm}8.9}$ & $35.6$ & $41.9_{\hspace{0.05cm}3.6}$ & $35.7$ & $67.3_{\hspace{0.05cm}17.2}$ & $26.4$ & 58.3\\
~  \UnOne & $74.8_{\hspace{0.05cm}15.6}$ & $50.0$ & $68.0_{\hspace{0.05cm}2.5}$ & $59.8$ & $54.8_{\hspace{0.05cm}6.2}$ & $47.1$ & $41.5_{\hspace{0.05cm}4.1}$ & $33.7$ & $82.3_{\hspace{0.05cm}4.5}$ & $64.9$ & 64.3\\
~  \UnBest-5 & $70.5_{\hspace{0.05cm}17.0}$ & $50.0$ & $66.2_{\hspace{0.05cm}3.4}$ & $54.6$ & $63.4_{\hspace{0.05cm}12.3}$ & $48.3$ & $45.7_{\hspace{0.05cm}4.7}$ & $33.6$ & $81.8_{\hspace{0.05cm}6.9}$ & $51.8$ & 65.5\\
~  \cellcolor{blue!25}\UnSimple & $\textbf{80.3}_{\hspace{0.05cm}12.8}$ & $50.0$ & $\textbf{69.0}_{\hspace{0.05cm}2.6}$ & $\textbf{61.5}$ & $\textbf{63.7}_{\hspace{0.05cm}11.7}$ & $\textbf{49.9}$ & $\textbf{48.1}_{\hspace{0.05cm}4.0}$ & $\textbf{39.2}$ & $\textbf{84.6}_{\hspace{0.05cm}3.1}$ & $\textbf{72.5}$ & $\textbf{69.2}$\\

\bottomrule
\end{tabular}}
\end{center}

\caption{Main results with different selection methods. The last column average accuracies across all tasks. Overall, the proposed methods \Simple\ and \Datamodels\ perform the best. Our method under the unlabeled setup (\UnSimple) even outperforms the \RandAll\ baseline that uses gold labels.}
\label{table:main_results}
\end{table*}

\section{Experiment}
\label{sec:exp}
\subsection{Setups}
\label{sec:exp_setup}
\paragraph{Tasks.} We experiment on 5 classification tasks: SST-2~\cite{socher2013recursive}, BoolQ~\cite{clark2019boolq}, Subj~\cite{Pang2004ASE}, Scicite~\cite{Cohan2019StructuralSF}, and AGNews~\cite{zhang2015character}.
We set the stable training subset size $E=20$ for all the tasks.
For binary classification tasks, we set $K=4$ and do not balance the classes in the prompts.
Thus, the collected $\Dicl$ for a binary task covers all $2^4$ label patterns, including prompts with all positive ($[1,1,1,1]$) and all negative ($[0,0,0,0]$) labels, allowing us to better understand the impact of label patterns on ICL.
For multiclass tasks (Scicite and AGNews), we balance the classes, sampling a training example per class to form the prompts.
Table~\ref{tab:tasks} in the appendix summarizes our setups.

\paragraph{Data Splits.} For all the tasks, we use class-balanced $\Dtrain$, $\Ddev$, and $\Dtest$ sampled from the original training set, as we do not have the gold labels of the original test set.
We choose $|\Dtrain| = 1000$ to ensure a diverse range of training examples for subset selection, and $|\Dtest| = 1000$ for reliable evaluation.
$\Ddev$ consists of 50 examples per class. 
All three sets are balanced, randomly sampled from the original training set, and do not overlap.

\paragraph{Models.} We run our main experiments with two LLMs: GPTJ-6B~\cite{gpt-j} and OPT-13B~\cite{zhang2022opt}.
More experiments on GPT-Neo-2.7B~\cite{gpt-neo} and OPT-6.7B can be found in Table~\ref{table:more_llm} in the appendix, where our methods also work well.

\subsection{Evaluation and Baselines}
\label{sec:eval_baseline}
Recall that our goal is to select a training subset $\mathcal{E} \subset \Dtrain$ that is more stable than $\Dtrain$.
To evaluate a selection method, we randomly sample 50 prompts from the selected subset $\mathcal{E}$, run ICL on the test set $\Dtest$, and report the average accuracy, standard deviation, and worst accuracy.

As shown in \citet{zhao2021calibrate},
when a prompt only contains examples of a single label,
LLMs are prone to
always predict that label on every test example.
Thus, in our main experiments (\cref{sec:main_results}), we ensure that every sampled prompt contains at least one example from every class.
In \cref{sec:single_label_results}, we separately investigate performance when the prompt contains only one label of binary tasks.
We split the selected subset $\mathcal{E}$ into two subsets, $\mathcal{E}_0$ and $\mathcal{E}_1$, where $\mathcal{E}_0$ only contains negative training examples and $\mathcal{E}_1$ only contains positive examples.
We then sample 50 prompts from $\mathcal{E}_0$ and $\mathcal{E}_1$, respectively.

Besides the two proposed selection methods, \Simple\ and \Datamodels, we design 5 baseline methods: \RandAll, \Calib, \RandSub, \One, and \Best.
\RandAll\ uses the entire training set as $\mathcal{E}$.
\Calib\ uses the same prompts as \RandAll, but with calibration to prevent LLMs biased toward certain labels, where we closely follow the implementation of \citet{zhao2021calibrate}.
\RandSub\ randomly selects a balanced training subset of $E=20$ examples.
\One\ first runs ICL with $K=1$, using each training example alone as the prompt, and then
scores the example by the corresponding ICL accuracy on $\Ddev$;
these scores are used in the same way as our main methods to select examples (\cref{sec:step3}).
\One\ tests if we can extrapolate ICL performance from $K=1$ to $K>1$.
\Best-5 and \Best-10 select the union of the examples from the top-$\{5,10\}$ prompts in $\Dicl$ with the highest dev set accuracy, where the subsets contain at most $K \times 5$ and $K \times 10$ examples, respectively.
Finally, we apply baselines and our \Simple\ method to the unlabeled setup (\cref{sec:problem_setup}), named with the \textsc{Un-} prefix.

\section{Results}
\label{sec:results}
\subsection{Main Results}
\label{sec:main_results}
\paragraph{The proposed methods outperform all baselines.} Table~\ref{table:main_results} shows the test set accuracy with different training subset selection methods.
Among methods without calibration, our \Simple\ and \Datamodels\ methods are the most \good, achieving substantially higher average and worst-case accuracy across all tasks, and lower variances on most tasks.
Compared with \Calib, our methods perform better in 8/10 setups.
Overall, \Simple\ and \Datamodels\ outperform the no-selection baseline \RandAll\ by 7.7\% and 6.3\% on average, respectively.

\paragraph{\Best\ is the strongest baseline.} Within the baselines, \RandAll\ and \RandSub\ have similar performance.
Applying calibration improves the worst accuracy on most tasks and the average accuracy on some tasks, but is not always beneficial, especially on Scicite.
\One\ outperforms \RandAll\ and \RandSub\ on SST-2 and BoolQ, but performs similarly on other tasks, indicating that we cannot extrapolate ICL behavior from $K=1$ to $K=3$ or $K=4$. 
\Best-5 and \Best-10 are the strongest baselines, performing especially well on SST-2, Subj, and Scicite, showing that the training examples that compose the highest-accuracy prompts are more stable than others.

\paragraph{Our method works without training set labels.} Randomly sampling prompts from the unlabeled training set (\UnAll) underperforms sampling from the original labeled training set (\RandAll), especially on SST-2 and AGNews.
This shows that gold labels do matter in ICL in general, in contrast with the findings of \citet{min2022rethinking}.
However, applying our selection method to the unlabeled training set (\UnSimple) surprisingly outperforms not only \UnAll\ but \RandAll\ (which uses correctly labeled examples), although some selected training examples actually have the wrong labels, 
implying that having gold-labeled prompts is not necessary for ICL.
Other baselines, \UnOne\ and \UnBest, outperform \UnAll\ but substantially underperform our method.
Overall, \UnSimple\ outperforms baselines \UnAll\ and \RandAll\ by 10.2\% and 5.7\% on average, respectively.

\paragraph{Does \UnSimple\ benfit from gold labels?}
We study the number of the stable training examples selected by \UnSimple\ that indeed have gold labels.
In most of the tasks, the numbers are much higher than the expected numbers by majority guess, where BoolQ is the exception with half of the selected examples having wrong labels.
We thus study if we can achieve even better results by correcting those selected examples that have wrong labels with their gold labels; the other correct examples in the subset remain unchanged.
After the label correction on BoolQ, the average and worst accuracy drops by 1.9\% and 4.5\% respectively on GPTJ, 0.4\% and 5.7\% respectively on OPT.
These results again suggest that on the one hand, ICL benefits from gold-labeled examples in most cases; on the other hand, some training examples with wrong labels can surprisingly achieve better performance.
The full results are in Table~\ref{table:unlabel} in the appendix.

Finally, we study the alternative that uses more shots in \ref{app:nshots}.
Table~\ref{table:maxshot} shows that using 4 curated examples (\Simple) can outperform $K=24, 16$ randomly sampled ones in SST-2 and AGNews.

\subsection{Single-Label Prompts}
\label{sec:single_label_results}
We now evaluate whether it is possible to achieve good accuracy while only using training examples of a single class in a prompt (See~\cref{sec:eval_baseline} for more details).
Table~\ref{table:single_label} compares the results of different methods.
First, the baselines \RandAll\ and \Best\ perform near chance in most cases, as the LLMs are biased by the prompts to predict the same label on every test example.
In contrast, single-label prompts sampled from the subsets of \Simple\ and \Datamodels\ substantially outperform majority guessing across all setups.
We conclude that the selected training examples are beneficial because they help LLMs understand the overall definition of the task.
Thus, even when used in single-label prompts, they can still give LLMs useful signal to perform the desired task.
\begin{table}[!t]
\begin{center}
\centering
\resizebox{1.0\columnwidth}{!}
{
\begin{tabular}{lcccc}
\toprule
      & \multicolumn{2}{c}{\textbf{SST-2}} & \multicolumn{2}{c}{\textbf{BoolQ}}\\

\cmidrule(lr){2-3}
\cmidrule(lr){4-5}
& [0,0,0,0] & [1,1,1,1] & [0,0,0,0] & [1,1,1,1]\\
\midrule
\emph{GPTJ-6B}\\
~  \RandAll & $51.7_{\hspace{0.05cm}1.7}$ & $52.6_{\hspace{0.05cm}3.1}$ & $52.8_{\hspace{0.05cm}1.6}$ & $50.7_{\hspace{0.05cm}1.0}$ \\
~  \Best & $52.1_{\hspace{0.05cm}2.1}$ & $56.0_{\hspace{0.05cm}3.7}$ & $54.2_{\hspace{0.05cm}1.9}$ & $51.8_{\hspace{0.05cm}1.6}$ \\
~  \Simple & $61.8_{\hspace{0.05cm}4.9}$ & $60.3_{\hspace{0.05cm}2.8}$ & $58.3_{\hspace{0.05cm}2.1}$ & $55.5_{\hspace{0.05cm}1.7}$ \\
~  \Datamodels & $\textbf{72.8}_{\hspace{0.05cm}4.9}$ & $\textbf{68.4}_{\hspace{0.05cm}4.9}$ & $\textbf{61.7}_{\hspace{0.05cm}1.6}$ & $\textbf{56.9}_{\hspace{0.05cm}2.2}$ \\

\midrule

\emph{OPT-13B}\\
~  \RandAll & $54.0_{\hspace{0.05cm}4.1}$ & $73.3_{\hspace{0.05cm}7.4}$ & $52.4_{\hspace{0.05cm}2.6}$ & $51.2_{\hspace{0.05cm}1.7}$ \\
~  \Best & $53.0_{\hspace{0.05cm}3.0}$ & $76.5_{\hspace{0.05cm}6.9}$ & $53.0_{\hspace{0.05cm}3.2}$ & $51.4_{\hspace{0.05cm}2.1}$ \\
~  \Simple & $\textbf{66.3}_{\hspace{0.05cm}3.7}$ & $81.0_{\hspace{0.05cm}4.6}$ & $65.3_{\hspace{0.05cm}2.9}$ & $53.7_{\hspace{0.05cm}1.9}$ \\
~  \Datamodels & $63.9_{\hspace{0.05cm}4.2}$ & $\textbf{84.5}_{\hspace{0.05cm}2.6}$ & $\textbf{69.2}_{\hspace{0.05cm}2.0}$ & $\textbf{60.1}_{\hspace{0.05cm}2.2}$ \\

\bottomrule
\end{tabular}}
\end{center}

\caption{Results of single-labeled prompts with different selection methods. Each prompt consists of 4 training examples with the same labels.}
\label{table:single_label}
\end{table}
\begin{table}[!t]
\begin{center}
\centering
\resizebox{0.95\columnwidth}{!}
{
\begin{tabular}{lcccc}
\toprule
      \textbf{OOD Tasks} & \multicolumn{2}{c}{\textbf{IMDb}} & \multicolumn{2}{c}{\textbf{BoolQ Cst.}}\\

\cmidrule(lr){2-3}
\cmidrule(lr){4-5}
 & Avg std & Min & Avg std & Min\\
\midrule
\emph{GPTJ-6B}\\
~  \RandAll & $86.5_{\hspace{0.05cm}5.7}$ & $63.6$ & $56.6_{\hspace{0.05cm}3.0}$ & $50.1$ \\
~  \Best & $87.2_{\hspace{0.05cm}5.2}$ & $63.0$ & $56.7_{\hspace{0.05cm}2.6}$ & $49.9$ \\
~  \Simple & $90.5_{\hspace{0.05cm}1.8}$ & $\textbf{84.8}$ & $\textbf{58.9}_{\hspace{0.05cm}1.7}$ & $\textbf{54.6}$ \\
~  \Datamodels & $\textbf{91.6}_{\hspace{0.05cm}1.5}$ & $84.0$ & $57.6_{\hspace{0.05cm}1.9}$ & $54.0$ \\


\midrule
\emph{OPT-13B}\\
~  \RandAll & $79.2_{\hspace{0.05cm}12.1}$ & $50.1$ & $59.8_{\hspace{0.05cm}2.9}$ & $51.6$ \\
~  \Best & $80.5_{\hspace{0.05cm}14.0}$ & $50.8$ & $60.3_{\hspace{0.05cm}3.5}$ & $51.0$ \\
~  \Simple & $83.5_{\hspace{0.05cm}10.8}$ & $54.6$ & $60.1_{\hspace{0.05cm}2.1}$ & $\textbf{56.7}$ \\
~  \Datamodels & $\textbf{84.1}_{\hspace{0.05cm}9.3}$ & $\textbf{58.9}$ & $\textbf{60.6}_{\hspace{0.05cm}3.3}$ & $54.3$ \\


\bottomrule
\end{tabular}}
\end{center}

\caption{Accuracy of IMDb and BoolQ Contrast Set, where the prompts consist of the selected SST-2 and BoolQ training examples, respectively.}
\label{table:ood}
\end{table}
\subsection{Out-of-Distribution Tasks}
We further evaluate on out-of-distribution (OOD) tasks, where there is a distribution shift between prompts and test data.
Specifically, we apply our selection methods on a source task, sampling 50 prompts from the selected subsets of the source task as done in the main experiments, and then evaluate on test data of a target task.
We use SST-2 and BoolQ as the source tasks, and IMDb \cite{imdb} and BoolQ Contrast Set \cite{gardner2020evaluating} as our target tasks, respectively.
Table~\ref{table:ood} shows that our \Simple\ and \Datamodels\ methods still outperform baselines on OOD tasks, especially on IMDb, implying that instead of simply overfitting the source tasks, the selected stable examples are indeed task-level examples that can generalize to OOD test data.

\begin{figure*}[!t]
\centering
\includegraphics[width=1.\linewidth]{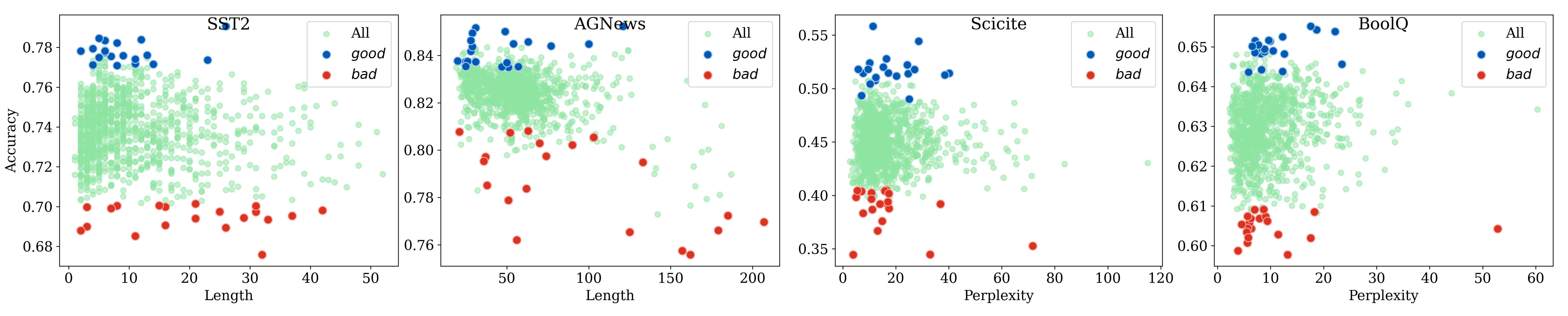}
\caption{Accuracy versus sequence length (left) and accuracy versus perplexity (right). Each dot corresponds to a training example. Examples in \emph{good} subsets are not outliers with abnormally long lengths or high perplexities.}
\label{fig:len_ppl_acc}
\end{figure*}
\section{Analysis}
\label{sec:analysis}
We further analyze what makes the selected training examples special along different dimensions: sequence length, perplexity, and diversity.
We compare \emph{good} with \emph{bad} training examples, where we use our \Simple\ method to identify a \emph{bad} (resp., \emph{good}) subset by selecting $E'$ training examples with the \emph{lowest} (resp., \emph{highest}) importance scores in each class (\cref{sec:step3}).
Please refer to Table~\ref{table:worst} in the appendix for the full results of the \emph{bad} subset.

\subsection{Sequence Length and Perplexity}
\label{sec:len_ppl}
In Figure~\ref{fig:len_ppl_acc}, we plot the accuracy against either sequence length or perplexity, where each dot corresponds to a training example.
Here, the accuracy (Y-axis) is the importance score 
 $\ssim$ assigned to each training example by \Simple\ in Eq.~\ref{eq:simple_score}, i.e., the average dev-set accuracy when that example is combined with random other training examples in a random order.
\paragraph{Sequence Length.}
The first two subfigures show that while the \emph{bad} examples (red dots) span across different sequence lengths, the \emph{good} examples (blue dots) do not cover the tail distribution of long sequences.
We observe little correlation between accuracy and sequence length across different tasks and LLMs (see more in Figure~\ref{fig:appendix_len2acc}), except for a slightly negative correlation when the sequence length is very long,
suggesting that abnormally long training examples can hurt ICL performance.
\paragraph{Perplexity.}
We calculate the perplexity of the inputs of training examples with respect to the same LLMs we run ICL on.
Figure~\ref{fig:len_ppl_acc} shows that \emph{good} examples are not outliers that have high perplexity.
This could suggest future work filter out examples that have extraordinarily high perplexity in the training set before running ICL, and could be combined with active learning for ICL~\cite{Zhang2022ActiveES,su2022selective}, as we only need the unlabeled inputs to calculate perplexity.
However, we observe no correlation between accuracy and perplexity across all the tasks and LLMs (Figure~\ref{fig:appendix_ppl2acc}), implying that using perplexity alone is not enough for identifying good training examples.
Our findings are inconsistent with concurrent work \cite{gonen2022demystifying}, which shows that lower prompt perplexity strongly correlates with better performance, probably because \citet{gonen2022demystifying} focus on perplexities under different instructions, while we focus on the differences between training inputs.

\subsection{Diversity}
\label{sec:diversity}
\begin{figure*}[!t]
\centering
\includegraphics[width=1.\linewidth]{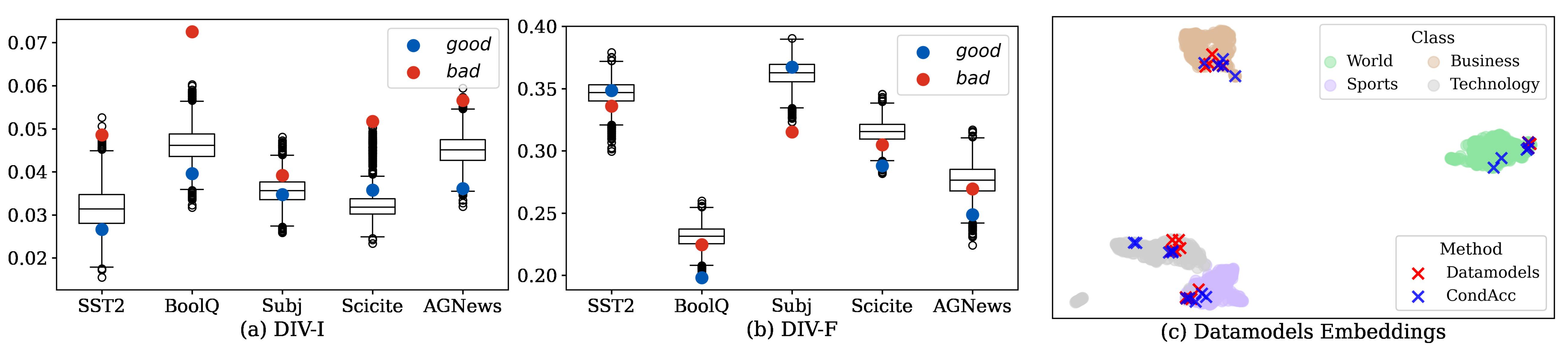}
\caption{Different ways to visualize the diversity of examples. (a) and (b) compare the diversity of the \emph{good} subset, \emph{bad} subset, and randomly sampled subsets (boxplot). For both DIV-I and DIV-F, a higher number means a subset is more diverse. Overall, \emph{good} subsets are no more diverse than random subsets. (c) visualizes the stable training examples selected by \Simple\ and \Datamodels\ methods in Datamodels embeddings space, where each dot is a training example in AGNews. Both methods choose tightly cluttered examples instead of diverse ones.}
\label{fig:diversity}
\end{figure*}

\paragraph{DIV-I and DIV-F.} We measure the diversity of a training subset with DIV-I \cite{yuan2020cold} and DIV-F \cite{zhdanov2019diverse} metrics, following prior work in active learning.
DIV-I measures diversity in raw text, while DIV-F measures diversity in a feature space. 
For DIV-F, we use SentBERT~\cite{Reimers2019SentenceBERTSE} to encode the inputs of training examples into sentence embeddings, following \citet{su2022selective}.
We compare the selected \emph{good} subset and \emph{bad} subset with 5000 randomly sampled subsets $\subset \Dtrain$, each containing $E'$ training examples per class.
Figure~\ref{fig:diversity} shows that \emph{good} subsets (blue dots) sometimes have low diversity scores in both metrics, especially on BoolQ and AGNews.
Overall, they do not seem to be more diverse than randomly sampled subsets.
Our findings are different from \citet{su2022selective}, which finds that diverse training subsets are better for prompt retrieval. 
We hypothesize that diversity matters more when retrieving similar training examples for each test input, but is less important when using a single fixed prompt. 
On the other hand, the \emph{bad} subsets have much higher DIV-I scores than random subsets across different tasks, because they often include examples with long sequence lengths (see Figure~\ref{fig:len_ppl_acc}), covering more distinct unigrams.
However, in the SentBERT feature space, the \emph{bad} subsets are often less diverse than random subsets.

\paragraph{Datamodels Embeddings.}
In \cref{sec:datamodel}, we learn a datamodel for each dev example in $\Ddev$.
Here, we concatenate the weights assigned on a training example learned by all the datamodels, creating an embedding $\in \mathbb{R}^{|\Ddev| \times K}$ for each training example. 
We then project the embeddings of the entire training set to a two-dimensional space with UMAP~\cite{mcinnes2018umap} for visualization.
Figure~\ref{fig:diversity} and Figure~\ref{fig:appendix_datamodel_embed} in the appendix show that both \Simple\ and \Datamodels\ choose tightly clustered sets of examples in the embedding space, instead of diverse ones scattering over the training set.
Moreover, the two methods actually select similar examples in the embedding space, although having very different scoring methods.

\subsection{Do LLMs find the same stable examples?}
\label{sec:xllm}
We further study if the identified stable subset examples are transferable across different LLMs.
Given two LLMs, we calculate the Pearson correlation coefficient between their example scores assigned by \Simple\ (\cref{sec:simple}) and the actual number of overlapped examples in the stable subsets of the two LLMs.
Table~\ref{table:cross_llms} shows the mixed results: some pairs have moderate correlation, especially when both LLMs are from the OPT family; however, we find little correlation between many pairs and there are only a few overlapped examples between the stable subsets identified by different LLMs.

Interestingly, when using the four stable SST2 examples shared by GPTJ-6B and OPT-13B as the prompt, evaluating all $4!$ permutations, we achieve very high test accuracy:  $88.6\pm 3.7$ on GPTJ, $89.8\pm3.0$ on OPT-6.7B, and $87.3 \pm 5.2$ on OPT-13B.
This may indicate that there exist successful factors of training examples shared among LLMs.
We include the four stable examples in Table~\ref{tab:xllm} and hope future work can discover what distinguishes them from other examples.
\begin{table}[!t]
\begin{center}
\centering
\resizebox{0.8\columnwidth}{!}
{
\begin{tabular}{lccc}
\toprule
& Corr & Overlap \\
\toprule
\textcolor{blue}{\emph{GPTJ-6B vs OPT-13B}}\\
~  \textbf{SST-2} & 0.15 & 4\\
~  \textbf{AGNews} & 0.46 & 1\\
~  \textbf{BoolQ} & 0.41 & 2\\
~  \textbf{Subj} & -0.03 & 2\\
~  \textbf{Scicite} & 0.02 & 2\\

\midrule
\textcolor{blue}{\emph{GPTJ-6B vs OPT-6.7B}}\\
~  \textbf{SST-2} & 0.27 & 3\\
~  \textbf{AGNews} & 0.08 & 1\\
\midrule
\textcolor{blue}{\emph{OPT-6.7B vs OPT-13B}}\\
~  \textbf{SST-2} & 0.76 & 3\\
~  \textbf{AGNews} & 0.42 & 1\\
\bottomrule
\end{tabular}}
\end{center}

\caption{Pearson correlation between the example scores ($\ssim$) of two LLMs and the number of overlapped examples in their stable subsets.}
\label{table:cross_llms}
\end{table}
\begin{table}[!t]
\begin{center}
\small
\begin{tabular}{p{0.9\columnwidth}}
\toprule
\textbf{Examples}\\
\midrule
\textbf{Review}: \textit{k-19 : the widowmaker is a great yarn.}\\
\textbf{Sentiment}: positive\\
\midrule
\textbf{Review}: \textit{spiffy animated feature}\\
\textbf{Sentiment}: positive\\
\midrule
\textbf{Review}: \textit{a plot cobbled together from largely flat and uncreative moments}\\
\textbf{Sentiment}: negative\\
\midrule
\textbf{Review}: \textit{has the thrown-together feel of a summer-camp talent show : hastily written, underrehearsed, arbitrarily plotted and}\\
\textbf{Sentiment}: negative\\

\bottomrule
\end{tabular}
\end{center}
\caption{The four stable subset examples shared by GPTJ-6B and OPT-13B in SST-2 dataset, where the last example is also selected by OPT-6.7B. All three models achieve high average accuracy and low variance across the $4!$ permutations of these four examples.}
\label{tab:xllm}
\end{table}
\section{Discussion and Related Work}
\label{sec:background}
\subsection{Prompt Retrieval}
The performance of ICL greatly depends on the choice of the prompt.
A common way to do prompt selection is to retrieve the top-$K$ similar training examples for each test input \cite{liu2021makes,rubin2021learning,su2022selective}, where the similarity is captured by sentence embeddings~\cite{Reimers2019SentenceBERTSE,robertson2009probabilistic}.
Such instance-dependent prompt retrieval is critical for semantic parsing tasks \cite{rubin2021learning}, as LLMs need to see relevant logical forms in the context to generate the appropriate predicates for the test example.
In this paper, we focus on classification tasks and identify task-level training examples that work for \emph{all} test examples, avoiding the retrieval process.

\subsection{The influence of in-context labels}
The proportions of labels appearing in context can greatly bias LLMs' predictions \cite{zhao2021calibrate}.
However, with careful prompt selection, \citet{Zhang2022ActiveES} finds that LLMs can perform well without observing the entire label space of a classification task in the prompt.
In \cref{sec:single_label_results}, we identify a subset, instead of just one prompt, of single-label examples that perform well.
On the other hand, the correctness of in-context labels may not matter as much, as \citet{min2022rethinking} find that randomly flipping them barely hurts performance.
\citet{Kim2022GroundTruthLM} re-examine the importance of gold labels, showing it varies largely across tasks and experimental settings; our unlabeled experiments also show varying importance of gold labels across different tasks.
Our method also identifies some training examples with wrong labels that can yield surprisingly good performance.

\subsection{Data Valuation}
 Given a learning algorithm trained on a training set to produce a predictor, data valuation quantifies the value of each training example to the predictor performance.
 Prior work includes influence functions~\cite{koh2017understanding}, Data Shapley~\cite{ghorbani2019data}, DVRL~\cite{yoon2020data}, TracIn~\cite{pruthi2020estimating}, and Datamodels~\cite{ilyas2022datamodels}.
 Our setup also aims to attribute the performance of a predictor (in our case, an LLM) to each training example (in our case, in-context examples).
Our \Simple\ method closely resembles Data Shapley, and we adapt \citet{ilyas2022datamodels} as our \Datamodels\ method.
However, the main difference is that we are doing $K$-shot ICL, where training examples are used as prompts, and there are no parameter updates to LLMs.

In concurrent and independent work, \citet{nguyen2023context} also propose methods based on Data Shapley and Datamodels to study the influence of training examples.
The main differences are: (1) we adapt Datamodels to consider the positions of training examples, while \citet{nguyen2023context} follow the original implementation and study example ordering with influence scores.
(2) They propose a Perplexity baseline that selects examples according to individual perplexity, while we use perplexity in analysis to study the correlation between example perplexity and their average performance. 
(3) \citet{nguyen2023context} explore a larger number of shots in ICL ($K \in [10,  52]$), while we assume a few-shot setting and fix $K \in \{3,4\}$.
Our findings on good training examples are consistent with each other: both papers find little correlation between performance and potential factors such as example length and perplexity.
In general, our work demonstrates the importance of data curation on in-context examples, even in the unlabeled and OOD scenarios, while \citet{nguyen2023context} focus more on developing influence-based example selection frameworks.
Taken together, two papers present a comprehensive view of data valuation for in-context learning.

\section{Conclusion}
We propose two methods to identify stable training subsets for in-context learning, achieving substantially higher average and worst-case accuracy across different setups.
Our \Simple\ method is intuitive and easy-to-implement, while our \Datamodels\ method provides informative weights that enable further analyses.
The success of our methods implies that when provided with the ``right'' training set (in our case, a subset of 20 examples to randomly sample prompts from), ICL could be far less sensitive to the choice of training examples and their orders in a prompt.
Our analyses on stable subsets find that they do not contain outliers with especially long sequence lengths or high perplexities, and are also no more diverse than random subsets of the training data.
We hope our work is a step towards developing guidelines for finding or writing better training examples.

\section*{Limitations}
The main limitation of our work is the high memory and computation cost. 
As both our methods estimate the importance of training
examples based on the prompt-performance statistics, we first need to run in-context learning on the dev set multiple times with different prompts in $\Dicl$.
Although ICL does not require any parameter updates, LLMs still require a large amount of memory footprint during inference, especially when the model size is large and the average sequence length is long.
For each setup, our $\Dicl$ contains around 50,000 prompts (see Table~\ref{tab:tasks}) and 50 dev examples per class, so the most expensive setup (running OPT-13B on BoolQ) costs more than 500 GPU hours on an RTXA6000 GPU.
Our preliminary study shows that the proposed methods need the statistics of at least 10,000 randomly sampled prompts to perform well.
Future work may use search algorithms instead of random sampling during data collection to reduce the number of prompts in $\Dicl$.
We also release our collected data of every setup in \url{https://github.com/terarachang/DataICL} to support future studies on ICL.

In this paper, we only study classification tasks, for the sake of easy evaluation.
Future work may study the influence of in-context examples in generation tasks under different evaluation metrics.
Due to hardware constraints, we do not study LLMs of sizes over 13B, and we fix the number of shots and prompt templates for simplicity.
In independent work, \citet{nguyen2023context} complement these limitations of our paper, showing that similar approaches work well on larger models and a diverse number of shots for in-context example selection.
Still, the influence of in-context examples for gigantic LLMs larger than 100B parameters has not been studied.
Due to emergent abilities of LLMs \cite{wei2022emergent}, it is unclear whether our methods and findings would still apply when prompting these gigantic LLMs.

\section*{Acknowledgements}

We thank Ameya Godbole, Johnny Wei, Wang Zhu, Albert Xu, Deqing Fu, Qinyuan Ye, the members of USC NLP, and
our anonymous reviewers for valuable feedback on the paper.
We thank Sameer Singh and Matt Gardner for helpful discussions.
We thank Jesse Thomason for his support.
This work was funded in part by gifts from Open Philanthropy and Google.

\bibliography{anthology,custom}
\bibliographystyle{acl_natbib}

\clearpage
\appendix
\section{Appendix}
\label{sec:appendix}
\subsection{Connection with Data Shapley}
\label{app:shapley}
Data Shapley has a valuation function $V(\DMsubset)$ for any subset of training examples. 
We define $V(\DMsubset)$ as the expected dev set ICL accuracy across all permutations of $\DMsubset$ if $|\DMsubset| = K$, and $V(\DMsubset)=0$ otherwise since we focus on $K$-shot learning.
Then, the Data Shapley value $\sshap(i)$ for the $i$-th training example $e_i = (x_i, y_i)$ is:
\begin{align*}
    \mathbb{E}_{ \{z_1, ..., z_{K-1}\}  \sim \binom{\Dtrain \setminus{e_i}}{K-1}} 
    \left[ V \bigl( \{z_1, ..., z_{K-1}, e_i \} \bigl) \right] \\
    - 
    \mathbb{E}_{ \{z_1, ..., z_{K}\}  \sim \binom{\Dtrain \setminus{e_i}}{K}} \left[ V \bigl( \{z_1, ..., z_{K}\} \bigl) \right].
    \label{eqn:shapley}
\end{align*}
We claim that $\sshap(i)$ is a monotonically increasing (in fact, affine) function of $\ssim(i)$, thus establishing a very tight connection between \Simple\ and Data Shapley. 

First, note that the first term is exactly equal to $\ssim(i)$, the expected conditional accuracy when $e_i$ occurs in a prompt, since $V$ returns the expected accuracy over all orders of $\{z_1, \dotsc, z_{K-1}, e_i\}$.

Similarly, the second term is the expected conditional accuracy when $e_i$ does not occur in a prompt.
Denote this quantity as $t(i)$,
and let $A$ denote the overall expected accuracy when randomly sampling a prompt.
Since the probability of choosing a given example to be in a prompt is exactly $\frac{K}{\Ntrain}$, we have
\begin{align*}
A &= \frac{K}{\Ntrain} \ssim(i) + \frac{\Ntrain - K}{\Ntrain} t(i) \\
t(i) &= \frac{\Ntrain}{\Ntrain - K} \left(A - \frac{K}{\Ntrain} \ssim(i) \right) \\
&= \frac{\Ntrain}{\Ntrain - K}A - \frac{K}{\Ntrain - K} \ssim(i).
\end{align*}

Now we can rewrite $\sshap(i)$ as follows:
\begin{align*}
&\sshap(i) = \ssim(i) - t(i) \\
&= \ssim(i) - 
\left(\frac{\Ntrain}{\Ntrain - K}A - \frac{K}{\Ntrain - K} \ssim(i)\right) \\
&= \frac{\Ntrain}{\Ntrain - K}A + \frac{\Ntrain}{\Ntrain - K} \ssim(i).
\end{align*}

Since $\Ntrain$, $K$, and $A$ are all constants that do not depend on $i$, and $\Ntrain, K > 0$, this establishes that $\sshap(i)$ is a monotonically increasing affine function of $\ssim(i)$.

\subsection{Relation to Prompt Tuning}
At test time, our setup is similar to Prompt Tuning~\cite{lester2021power}, where a fixed prompt for a task is prepended to the test inputs.
In our case, a prompt is a sequence of training examples randomly drawn from the selected subset (fixed for all test examples).
In Prompt Tuning, it is a set of continuous embeddings, called a \emph{soft prompt}, learned through backpropagation.

At training time, however, Prompt Tuning needs to backpropagate through the LLM and thus is more memory-expensive and tends to suffer from training instability~\cite{Asai2022AttentionalMO}.
In comparison, our method finds good prompts without accessing the LLM's parameters, which is a realistic setup as many LLMs only provide API access.

\begin{table}[!t]
\begin{center}
\centering
{
\begin{tabular}{lcccc}
\toprule
      & \multicolumn{2}{c}{\textbf{GPTJ-6B}} & \multicolumn{2}{c}{\textbf{OPT-13B}}\\

\cmidrule(lr){2-3}
\cmidrule(lr){4-5}
& $L_1$ & Corr & $L_1$ & Corr\\
\midrule

\textbf{SST2} & 0.133 & 0.962 & 0.264 & 0.930 \\
\textbf{Boolq} & 0.147 & 0.941 & 0.167 & 0.937\\
\textbf{Subj} & 0.269 & 0.946 & 0.260 & 0.949\\
\textbf{Scicite} & 0.088 & 0.937 & 0.151 & 0.938\\
\textbf{AGNews} & 0.296 & 0.891 & 0.340 & 0.840\\

\bottomrule
\end{tabular}}
\end{center}

\caption{Test results of datamodels. Our datamodels can accurately predict LLMs' outcomes on unseen prompts, having high correlation and low $L_1$ distance to the ground-truth outcomes.}
\label{table:dm_test}
\end{table}
\begin{table*}[!t]
\begin{center}
\resizebox{2.\columnwidth}{!}{%
\begin{tabular}{llc}
\toprule
\textbf{Task} &  \textbf{Example} & \textbf{Label Mapping}\\
\bottomrule
SST-2 & \begin{tabular}[c]{@{}l@{}}Review: {contains no wit , only labored gags}\\Sentiment: negative\end{tabular} & negative/positive \\
\midrule

BoolQ & \begin{tabular}[c]{@{}l@{}}Exercise: read the text and answer the question by yes or no.\\\\{Good Samaritan laws offer legal protection to people who give reasonable assistance...}\\Question: do good samaritan laws protect those who help at an accident? yes \end{tabular} & no/yes \\
\midrule

Subj & \begin{tabular}[c]{@{}l@{}}Input: {the tucks have a secret , they 're immortal .}\\Type: objective\end{tabular} & objective/subjective \\
\midrule

Scicite & \begin{tabular}[c]{@{}l@{}}Is the following citation from a scientific paper describing a method, a result, or background?\\{However, how frataxin interacts with the Fe-S cluster biosynthesis components...}\\Answer: background \end{tabular} & \makecell{method/result/ \\background} \\
\midrule

MNLI & \begin{tabular}[c]{@{}l@{}} \makecell[l]{"yeah well you're a student right" Based on the previous passage, is it true that "Well you're \\ a mechanics student right"? Yes, no, or maybe? maybe}\end{tabular} & yes/maybe/no \\
\midrule

AGNews & \begin{tabular}[c]{@{}l@{}}Article: {Wall St. Bears Claw Back Into the Black (Reuters) Reuters - Short-sellers...}\\Answer: Business\end{tabular} & \makecell{World/Sports/\\Business/Technology} \\

\bottomrule
\end{tabular}
}
\end{center}
\caption{Our templates and label mappings in different tasks. For simplicity, all the label words we use consist of a single token, so we can easily get the probability of each label.}
\label{tab:appendix_prompt}
\end{table*}
\subsection{Evaluating Datamodels}
\label{app:dm_test}
\begin{figure*}[h!]
\centering
\begin{subfigure}[b]{1\textwidth}
   \includegraphics[width=1\linewidth]{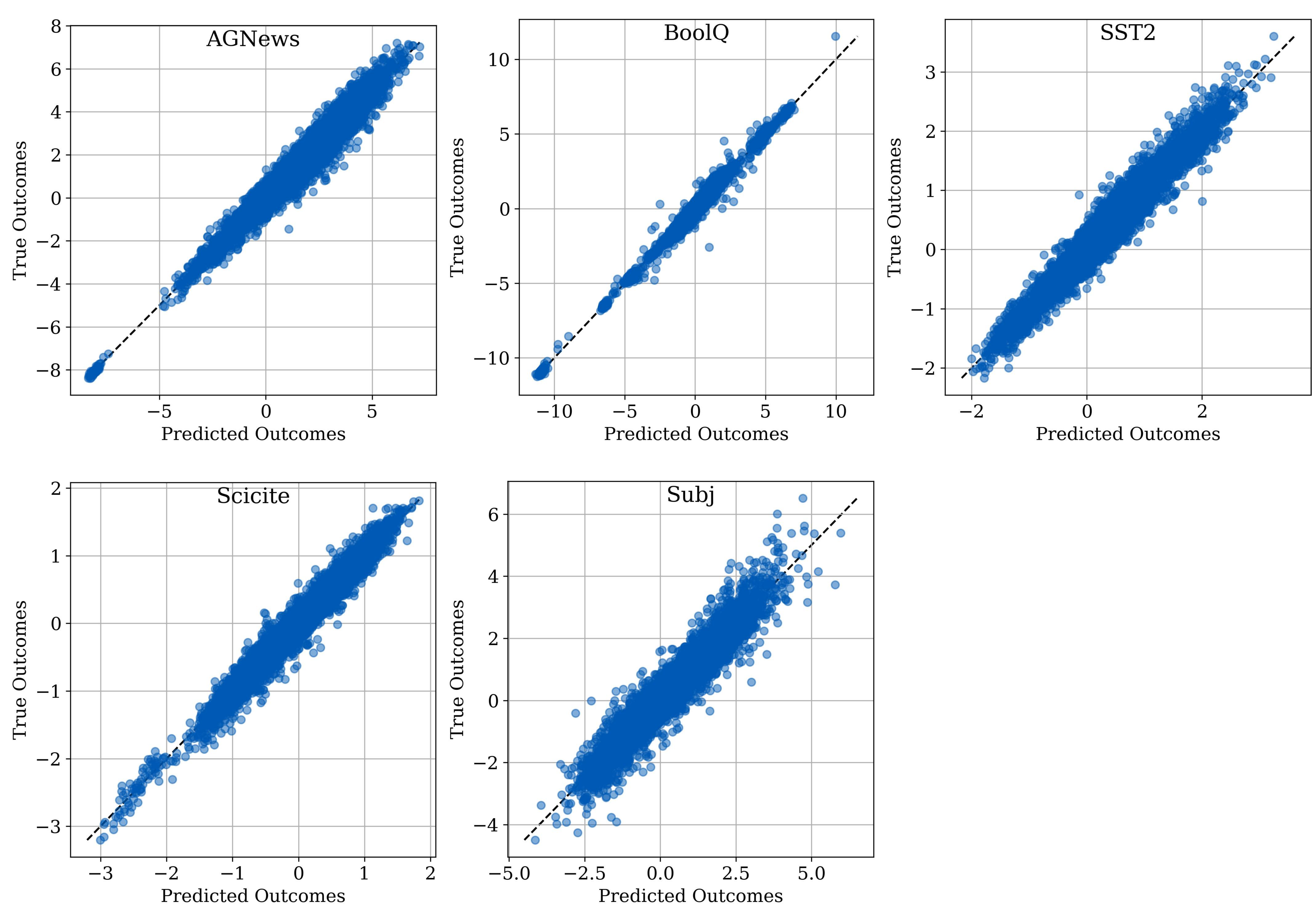}
   \caption{GPTJ-6B}
\end{subfigure}

\begin{subfigure}[b]{1\textwidth}
   \includegraphics[width=1\linewidth]{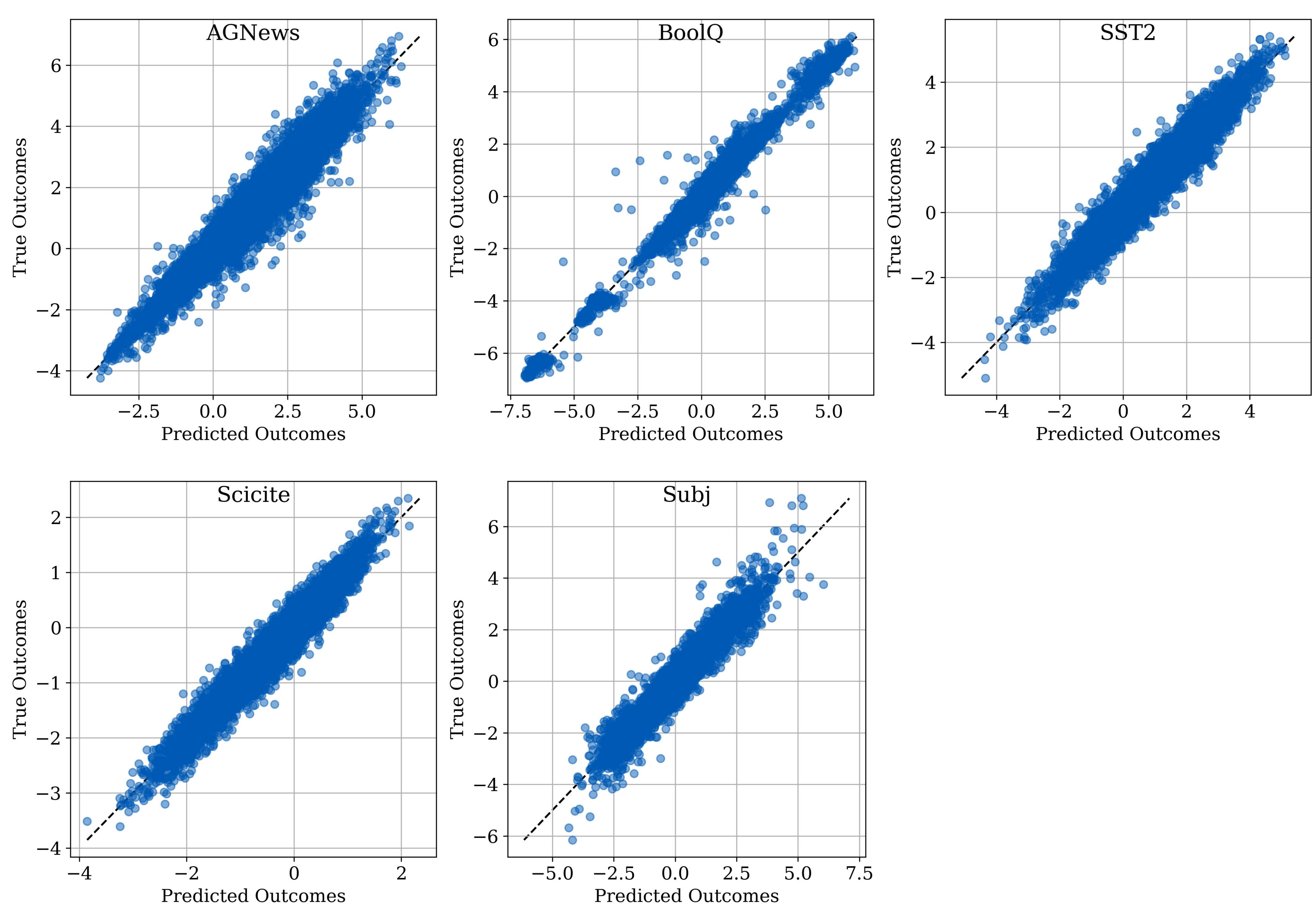}
   \caption{OPT-13B}
\end{subfigure}
\caption{The ground-truth outcomes of an LLM versus predicted outcomes of datamodels on the test set of datamodels, which contains a set of newly sampled prompts with unseen combinations of training examples. The high correlations show that our datamodels can make accurate predictions.}
\label{fig:appendix_dm_test}
\end{figure*}

Recall that given a prompt $\Prompt$ and a dev example $(\DevInp, \DevOut)$, a datamodel learns to approximate an LLM's outcome of $\DevInp$ (\cref{sec:datamodel}).
To evaluate how accurate our datamodels can approximate an LLM, we create a test set for datamodels $\mathcal{D}_{\text{DM}}$, consisting of 5000 pairs of newly sampled $\Prompt$ and the LLM's ground-truth outcomes of every dev example, where our sampling assures that every $\Prompt$ is made up of an unseen combination of training examples.
We evaluate the learned datamodels on $\mathcal{D}_{\text{DM}}$, calculating the correlation and $L_1$ distance between the predicted outcomes and the ground-truth outcomes.
More specifically, each datamodel yields 5000 outcomes, which we calculate the Pearson correlation coefficient with the ground-truth outcomes of the associated dev example.
We report the average correlation and $L_1$ distance over all datamodels in Table~\ref{table:dm_test}.
We also randomly sample 10,000 outcomes across all datamodels to visualize the ground truths against predictions in Figure~\ref{fig:appendix_dm_test}.

Overall, our datamodels can accurately approximate LLMs' outcomes across different tasks and LLMs.
As our datamodels only consider two simple features, the existence of a training example and its index in $\Prompt$, accurate test predictions may imply that these two features have a dominating effect on ICL.

\subsection{Training Details of Datamodels}
As the pattern of in-context labels (e.g., [0,0,0,0], [0,0,0,1], [1,0,0,1]) has a great impact on LLMs' predictions, for each label pattern, we train a set of $|\Ddev|$ datamodels.
Specifically, we apply two-phase training:
in the first phase, we train on all data with shared weights.
In the second phase, we first bucket prompts in $\Dicl$ by their label patterns. 
Initializing from the weights learned in the first phase,
we then separately fine-tune a set of datamodels for each label pattern.
For example, for a binary task with 4-shot learning, there are $2^4=16$ label patterns; thus, we have $16$ sets of datamodels after the second-phase training, namely, $16 \times |\Ddev|$ datamodels in total.
We find that having two-phase training leads to more accurate predictions of LLMs' outcomes in \cref{app:dm_test}.
Thus, when assigning scores for training examples, we aggregate all sets of datamodels to obtain $\boldsymbol{s}_{dm}$.
When creating datamodel embeddings (\cref{sec:analysis}), we use the unified weights learned by the first phase.

\subsection{Implementation Details}
We use PyTorch and Huggingface transformers to implement in-context learning on GPT-Neo-2.7B~\cite{gpt-neo}, GPTJ-6B~\cite{gpt-j}, OPT-6.7B~\cite{zhang2022opt}, and OPT-13B.
We run all our evaluations on a single RTXA6000 GPU (48GB).
Most of our experiments can also be run on an RTX3090 GPU (24GB), except that OPT-13B model requires a GPU with larger memory.

Our data collection on $\Dicl$ costs hundreds of GPU hours on an RTXA6000.
Once we finish the collection, our \Datamodels\ method only takes about 5 seconds to train a datamodel on an i7-10700 CPU, and our \Simple\ method does not involve any training, but simply calculates accuracy. 

Table~\ref{tab:appendix_prompt} shows our task templates and label mappings, where we closely follow \citet{bach2022promptsource,lu2021fantastically}.
Table~\ref{tab:tasks} summarizes our experimental setups on different tasks. 

\subsection{More Experiments}
Table~\ref{table:more_llm} shows experiments on more LLMs and the MNLI task, where we evaluate on test data using 50 sampled prompts, as done in the main experiments (\cref{sec:eval_baseline}).
Since ICL performs poorly on MNLI (majority: 33.3\%) in both prior work~\cite{su2022selective} and our results in Table~\ref{table:more_llm}, we do not experiment more on this task. 
Overall, our methods \Simple\ and \Datamodels\ substantially outperform other 4 baselines on all setups.

\subsection{Why not Instruction-Finetuned LLMs?}
The massive multitask learning in instruction-tuning leads to leakage in the datasets we evaluate on. For example, T0~\cite{sanh2021multitask}, FLAN-T5~\cite{chung2022scaling}, and OPT-IML~\cite{iyer2022opt} are all trained on several tasks in our paper.

\begin{table*}
\renewcommand{\arraystretch}{1.5}
\begin{center}
\centering
\resizebox{2.0\columnwidth}{!}{%
\begin{tabular}{lccccccccc}
\toprule
Task &  $N_{\text{class}}$ & Bal. & $K$ & $|\Dtrain^{L}|$ & $|\Dtrain^{U}|$ & $\text{Permut}^{L}$ & $\text{Permut}^{U}$ & $|\Dicl^{L}|$ & $|\Dicl^{U}|$\\
\midrule
SST-2 & 2 & N & 4 & 1000 & 2000 & ${1000 \choose 4} \times 4! $ & ${2000 \choose 4}\times 4! $ & 100,000 & 50,000\\
BoolQ & 2 & N & 4 & 1000 & 2000 & ${1000 \choose 4} \times 4! $ & ${2000 \choose 4}\times 4! $ & 100,000 & 50,000\\
Subj & 2 & N & 4 & 1000 & 2000 & ${1000 \choose 4} \times 4! $ & ${2000 \choose 4}\times 4! $ & 100,000 & 50,000\\
\midrule
Scicite & 3 & Y & 3 & 999 & 2997 & $(333)^3 \times 3!$ & $(999)^3 \times 3!$ & 40,000 & 50,000\\
AGNews & 4 & Y & 4 & 1000 & 4000 & $(250)^4 \times 4!$ & $(1000)^4 \times 3!$ & 40,000 & 50,000\\

\bottomrule
\end{tabular}
}
\end{center}
\caption{Setups on different tasks. (1) We balance the classes in the prompts (Bal.) for multiclass tasks. (2) To create the unlabeled training set $\Dtrain^{U}$, we pair each input with every possible label; therefore, $\Dtrain^{U}$ is $N_{\text{class}}$ times larger than the gold-labeled training set $\Dtrain^{L}$. (3) $\text{Permut}^{L}$ and $\text{Permut}^{U}$ denote the total number of possible permutations of training examples for $K$-shot ICL in the labeled and unlabeled setups, respectively. (4) $|\Dicl^{L}|$ and $|\Dicl^{U}|$ denote the number of the prompts we collect in the labeled and unlabeled setups, respectively.}
\label{tab:tasks}
\end{table*}
\begin{table*}[!t]

\centering
\begin{tabular}{ cc }   

    \begin{minipage}{.55\linewidth}
        \resizebox{1\columnwidth}{!}{
        \begin{tabular}{lcccc}
        \toprule
              & \multicolumn{2}{c}{\textbf{SST-2}} & \multicolumn{2}{c}{\textbf{AGNews}}\\
        
        \cmidrule(lr){2-3}
        \cmidrule(lr){4-5}
        & Avg std & Min & Avg std & Min\\
        \midrule
        \emph{GPT-Noe-2.7B}\\
        ~  \RandAll & $64.5_{\hspace{0.05cm}13.0}$ & $50.0$ & $74.8_{\hspace{0.05cm}5.8}$ & $61.8$ \\
        ~  \RandSub & $65.2_{\hspace{0.05cm}12.8}$ & $50.0$ & $74.3_{\hspace{0.05cm}6.8}$ & $56.2$ \\
        ~  \One & $66.1_{\hspace{0.05cm}12.8}$ & $50.0$ & $78.3_{\hspace{0.05cm}4.4}$ & $64.9$ \\
        ~  \Best-5 & $64.1_{\hspace{0.05cm}12.7}$ & $50.0$ & $79.9_{\hspace{0.05cm}3.4}$ & $71.1$ \\
        ~  \cellcolor{blue!25}\Simple & $\textbf{76.5}_{\hspace{0.05cm}10.5}$ & $\textbf{52.4}$ & $82.3_{\hspace{0.05cm}2.2}$ & $77.4$ \\
        ~  \cellcolor{blue!25}\Datamodels & $72.6_{\hspace{0.05cm}14.2}$ & $50.4$ & $\textbf{83.5}_{\hspace{0.05cm}1.4}$ & $\textbf{80.0}$ \\
        
        \midrule
        \emph{OPT-6.7B}\\
        ~  \RandAll & $76.8_{\hspace{0.05cm}11.8}$ & $52.4$ & $67.9_{\hspace{0.05cm}15.8}$ & $26.0$ \\
        ~  \RandSub & $72.6_{\hspace{0.05cm}12.7}$ & $50.2$ & $66.1_{\hspace{0.05cm}14.7}$ & $27.6$ \\
        ~  \One & $84.7_{\hspace{0.05cm}6.6}$ & $65.7$ & $76.3_{\hspace{0.05cm}7.8}$ & $56.7$ \\
        ~  \Best-5 & $78.8_{\hspace{0.05cm}10.9}$ & $50.4$ & $78.2_{\hspace{0.05cm}9.7}$ & $30.8$ \\
        ~  \cellcolor{blue!25}\Simple & $\textbf{88.2}_{\hspace{0.05cm}5.8}$ & $59.4$ & $83.2_{\hspace{0.05cm}4.3}$ & $67.2$ \\
        ~  \cellcolor{blue!25}\Datamodels & $87.4_{\hspace{0.05cm}5.3}$ & $\textbf{71.7}$ & $\textbf{84.2}_{\hspace{0.05cm}3.0}$ & $\textbf{74.0}$ \\
        \bottomrule
        \end{tabular}}
    \end{minipage} &
    
    
    \begin{minipage}{.55\linewidth}
        \resizebox{0.66\columnwidth}{!}{
        \begin{tabular}{lcc}
        \toprule 
        & \multicolumn{2}{c}{\textbf{MNLI}}\\
              
        \cmidrule(lr){2-3}
        & Avg std & Min\\
        \midrule
        \emph{GPTJ-6B}\\
        ~  \RandAll & $43.8_{\hspace{0.05cm}2.9}$ & $35.8$ \\
        ~  \RandSub & $44.1_{\hspace{0.05cm}2.7}$ & $33.8$ \\
        ~  \One & $41.7_{\hspace{0.05cm}4.3}$ & $33.5$ \\
        ~  \Best-5 & $44.7_{\hspace{0.05cm}2.5}$ & $38.0$ \\
        ~  \cellcolor{blue!25}\Simple & $\textbf{45.6}_{\hspace{0.05cm}1.8}$ & $40.0$ \\
        ~  \cellcolor{blue!25}\Datamodels & $44.9_{\hspace{0.05cm}1.7}$ & $\textbf{40.3}$ \\
        \bottomrule
        \end{tabular}}
    \end{minipage}

\end{tabular}

\caption{More results of GPT-Neo-2.7B and OPT-6.7B (left) and GPTJ-6B on MNLI task (right). Overall, the proposed subset selection methods \Simple\ and \Datamodels\ significantly outperform other baselines.}
\label{table:more_llm}
\end{table*}

\begin{table}[!t]

\centering
{
\begin{tabular}{lcccc}
\toprule & \textbf{GPTJ-6B} & \textbf{OPT-13B} & \textbf{Majority}\\

\midrule
\textbf{SST-2} & 20 & 19 & 10 \\
\textbf{BoolQ} & 11 & 10 & 10 \\
\textbf{Subj} & 20 & 20 & 10 \\
\textbf{Scicite} & 16 & 11 & 6.6\\
\textbf{AGNews} & 18 & 13 & 5 \\

\bottomrule
\end{tabular}}

\caption{The number of gold-labeled training examples identified by \UnSimple\ in the unlabeled setup. The subset size $E=20$ for all tasks.}
\label{table:unlabel}
\end{table}
\begin{table}

\centering
{
\begin{tabular}{lcccc}
\toprule & \textbf{GPTJ-6B} & \textbf{OPT-13B} & \textbf{Majority}\\

\midrule
\textbf{SST-2} & $66.0_{\hspace{0.05cm}12.3}$ & $55.3_{\hspace{0.05cm}10.6}$ & $50.0$\\
\textbf{BoolQ} & $50.5_{\hspace{0.05cm}3.4}$ & $55.6_{\hspace{0.05cm}5.8}$ & $50.0$\\
\textbf{Subj} & $51.8_{\hspace{0.05cm}4.2}$ & $50.6_{\hspace{0.05cm}1.8}$ & $50.0$\\
\textbf{Scicite} & $33.9_{\hspace{0.05cm}1.2}$ & $36.2_{\hspace{0.05cm}2.6}$ & $33.3$\\
\textbf{AGNews} & $60.8_{\hspace{0.05cm}9.4}$ & $54.6_{\hspace{0.05cm}10.1}$ & $25.0$\\

\bottomrule
\end{tabular}}

\caption{Results of the \emph{bad} training subsets, which consist of examples of the lowest scores assigned by \Simple method.}
\label{table:worst}
\end{table}
\clearpage
\clearpage

\subsection{Larger Number of Shots}
\label{app:nshots}
We further compare our methods with an alternative that takes as many labeled, balanced, in-context examples as the context window can fit, named \MaxShot.
Similar to the \RandAll\ baseline, \MaxShot\ samples 50 prompts from the entire training set, each containing $K$ training examples.
The differences are that \MaxShot\ uses a much larger $K$ and balances the classes in the prompt for binary tasks as well.
Here, our LLM is GPTJ-6B, which has a context window of 2048 tokens.

Table~\ref{table:maxshot} shows the number of shots $K$ for each task and compares the average test set accuracy of \MaxShot\ with the ones of \RandAll\ and \Simple\ in Table~\ref{table:main_results}.
The $\Delta_\text{All}$ column shows that using $K \in [8,  24]$ examples in the prompt substantially improves over only 3 or 4 examples in most tasks, except for AGNews.
However, \MaxShot\ only outperforms \Simple\ on two out of five tasks ($\Delta_\text{Our}$, blue).
This shows the advantages of curated training examples over randomly sampled ones.
\begin{table}[!h]

\centering
\resizebox{1.0\columnwidth}{!}
{
\begin{tabular}{lcccc}
\toprule & $K$ & Avg std & $\Delta_\text{All}$ & $\Delta_\text{Our}$\\

\midrule
\textbf{SST-2} & 24 &$85.8_{\hspace{0.05cm}5.5}$ &\textcolor{blue}{8.0} &\textcolor{red}{-0.9} \\
\textbf{Subj} & 24 &$77.1_{\hspace{0.05cm}8.2}$ &\textcolor{blue}{17.2} &\textcolor{blue}{6.6} \\
\textbf{BoolQ} & 8 &$63.6_{\hspace{0.05cm}1.8}$ &\textcolor{blue}{2.7} &\textcolor{red}{-1.5} \\
\textbf{Scicite} & 12 &$57.1_{\hspace{0.05cm}6.2}$ &\textcolor{blue}{13.2} &\textcolor{blue}{4.8} \\
\textbf{AGNews} & 16 &$82.3_{\hspace{0.05cm}4.8}$ &\textcolor{red}{-1.2} &\textcolor{red}{-5.0} \\

\bottomrule
\end{tabular}}

\caption{Test results of \MaxShot\ on GPTJ, where $\Delta_\text{All}$ and $\Delta_\text{Our}$ show its improvements of average accuracy over \RandAll\ and \Simple, respectively (both only use $K=3, 4$ training examples).}
\label{table:maxshot}
\end{table}
\clearpage
\begin{figure*}[h!]
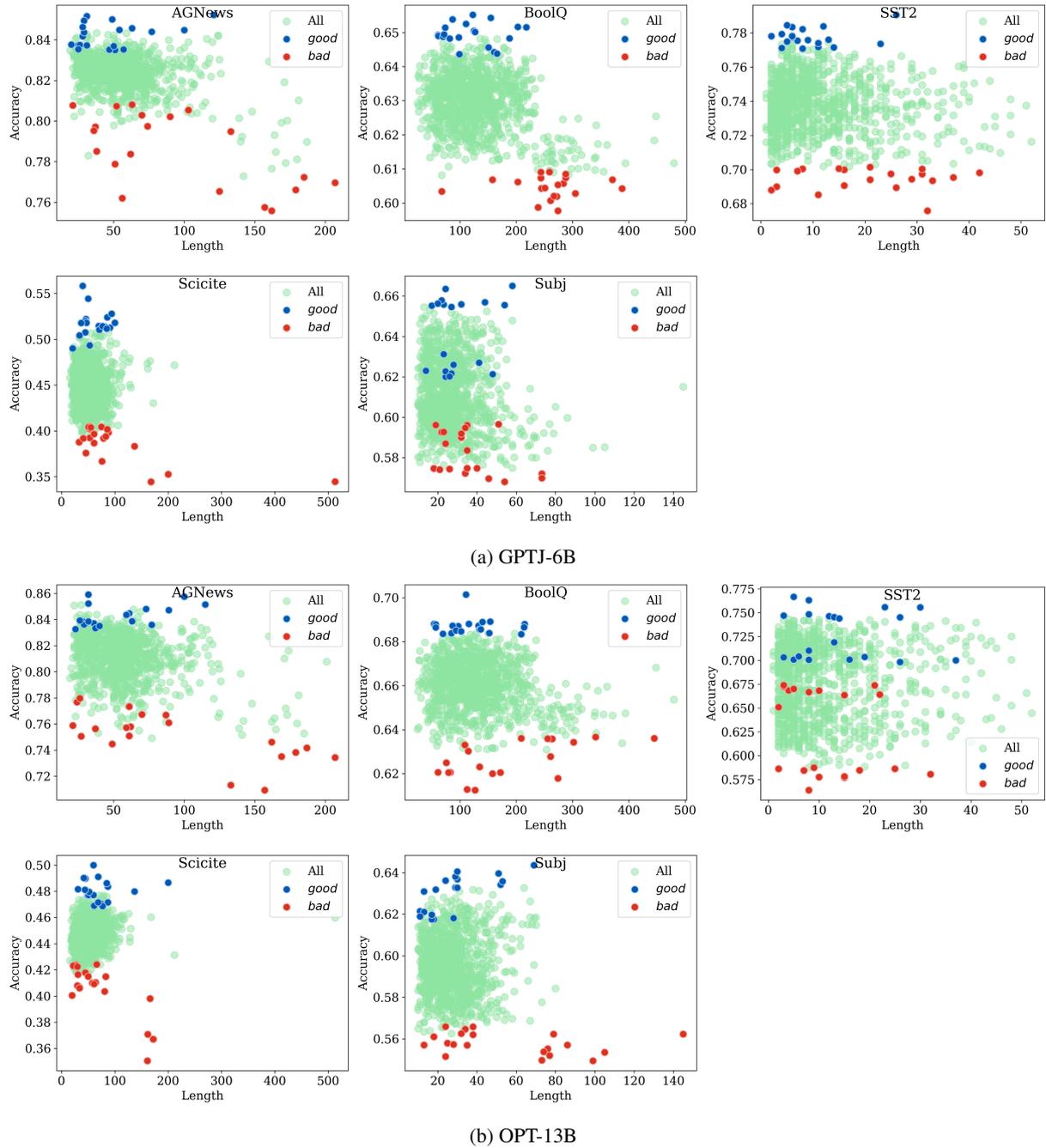

\centering
\begin{subfigure}[b]{1\textwidth}
   \includegraphics[width=1\linewidth]{Figs/all-Len2Acc-GPTJ.pdf}
   \caption{GPTJ-6B}
\end{subfigure}

\begin{subfigure}[b]{1\textwidth}
   \includegraphics[width=1\linewidth]{Figs/all-Len2Acc-OPT.pdf}
   \caption{OPT-13B}
\end{subfigure}
\caption{The accuracy versus sequence length across different tasks. Each dot corresponds to a training example. Note that we select the top $E^{\prime}$ examples per class. As a class may have much lower average accuracy than others, the \emph{good} (resp., \emph{bad}) examples may not be the examples with the globally highest (resp., lowest) accuracy.}
\label{fig:appendix_len2acc}
\end{figure*}
\begin{figure*}[h!]
\centering
\begin{subfigure}[b]{1\textwidth}
   \includegraphics[width=1\linewidth]{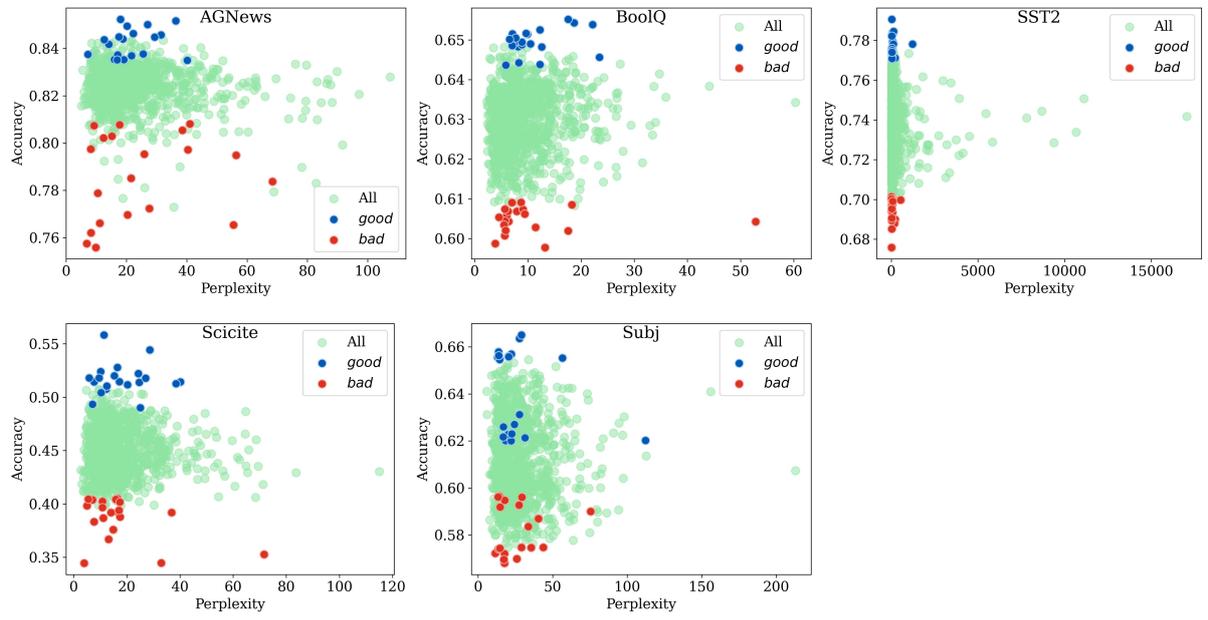}
   \caption{GPTJ-6B}
\end{subfigure}

\begin{subfigure}[b]{1\textwidth}
   \includegraphics[width=1\linewidth]{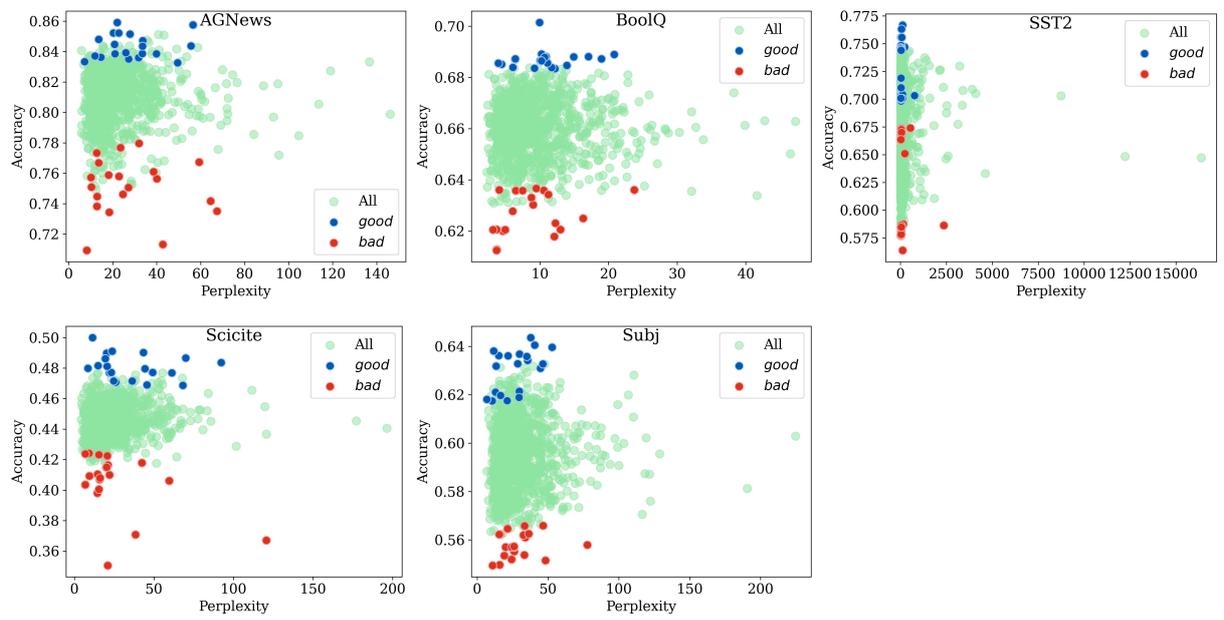}
   \caption{OPT-13B}
\end{subfigure}
\caption{The accuracy versus perplexity across different tasks. Each dot corresponds to a training example. We do not observe any correlation between perplexity and accuracy. Note that we select the top $E^{\prime}$ examples per class. As a class may have much lower average accuracy than others, the \emph{good} (resp., \emph{bad}) examples may not be the examples with the globally highest (resp., lowest) accuracy.}
\label{fig:appendix_ppl2acc}
\end{figure*}
\begin{figure*}[!t]
\centering
\includegraphics[width=0.85\linewidth]{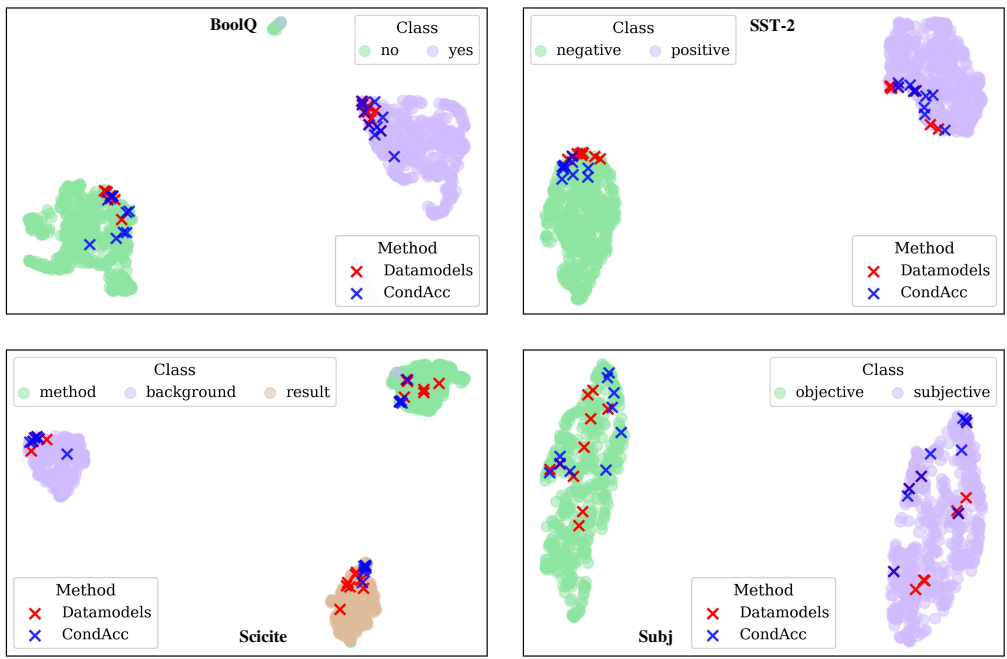}
\caption{Visualizing the \emph{good} training examples selected by \Simple\ and \Datamodels\ in datamodels embedding space across different tasks with GPTJ. Each dot is a training example, where datamodels spontaneously learn to encode class information in the embeddings.}
\label{fig:appendix_datamodel_embed}
\end{figure*}

\end{document}